\documentclass[sigconf,screen]{acmart}

\acmConference[ICSE 2024]{46th International Conference on Software Engineering}{April 2024}{Lisbon, Portugal}

\usepackage{xspace}
\usepackage{url}
\usepackage{color,xcolor}

\usepackage{multirow}
\usepackage{graphicx}

\usepackage[framemethod=tikz]{mdframed}
\newcounter{finding}
\newcommand{\finding}[1]{\refstepcounter{finding}
  \vspace{2.3mm}
 \begin{mdframed}[linecolor=gray,roundcorner=12pt,backgroundcolor=gray!15,linewidth=3pt,innerleftmargin=2pt, leftmargin=0cm,rightmargin=0cm,topline=false,bottomline=false,rightline = false]
  \textbf{Finding \arabic{finding}:} #1
 \end{mdframed}
 \vspace{2.3mm}
}
\usepackage{enumitem}

\usepackage{makecell,colortbl}

\copyrightyear{2024}
\acmYear{2024}
\acmConference[ICSE '24]{2024 IEEE/ACM 46th International Conference on Software Engineering}{April 14--20, 2024}{Lisbon, Portugal}
\acmBooktitle{2024 IEEE/ACM 46th International Conference on Software Engineering (ICSE '24), April 14--20, 2024, Lisbon, Portugal}

\begin{document}

\title[Fairness Improvement with Multiple Protected Attributes: How Far Are We?]{Fairness Improvement with Multiple Protected Attributes: \\How Far Are We?}

\author{Zhenpeng Chen}
\email{zp.chen@ucl.ac.uk}
\affiliation{%
  \institution{University College London}
  \city{London}
  \country{United Kingdom}
}

\author{Jie M. Zhang}\authornote{Corresponding author}
\email{jie.zhang@kcl.ac.uk}
\affiliation{%
  \institution{King's College London}
  \city{London}
  \country{United Kingdom}
}

\author{Federica Sarro}
\email{f.sarro@ucl.ac.uk}
\affiliation{%
  \institution{University College London}
  \city{London}
  \country{United Kingdom}
}

\author{Mark Harman}
\email{mark.harman@ucl.ac.uk}
\affiliation{%
  \institution{University College London}
  \city{London}
  \country{United Kingdom}
}

\renewcommand{\shortauthors}{Zhenpeng Chen, Jie M. Zhang, Federica Sarro, Mark Harman}

\begin{abstract}
Existing research mostly improves the fairness of Machine Learning~(ML) software regarding a single protected attribute at a time, but this is unrealistic given that many users have multiple protected attributes. This paper conducts an extensive study of fairness improvement regarding multiple protected attributes, covering 11 state-of-the-art fairness improvement methods. We analyze the effectiveness of these methods with different datasets, metrics, and ML models when considering multiple protected attributes. The results reveal that improving fairness for a single protected attribute can largely decrease fairness regarding unconsidered protected attributes. This decrease is observed in up to 88.3\% of scenarios (57.5\% on average). More surprisingly, we find little difference in accuracy loss when considering single and multiple protected attributes, indicating that accuracy can be maintained in the multiple-attribute paradigm. However, the effect on F1-score when handling two protected attributes is about twice that of a single attribute. This has important implications for future fairness research: reporting only accuracy as the ML performance metric, which is currently common in the literature, is inadequate. 
\end{abstract}



\maketitle

\section{Introduction}

Machine Learning (ML) software is being increasingly applied to assist decision-making in social-critical scenarios.
This has raised surging concerns on the fairness of such software \cite{csurMehrabiMSLG21}. Indeed, ML software frequently exhibits unfair behaviors related to protected attributes such as sex \cite{faircase,faircase2} and race \cite{faircase1,corrabs230802935}.
Unfair behaviors may compromise the benefits of historically disadvantaged groups, and lead to consequences for Software Engineering (SE) if and when the software is found to contravene laws against discrimination~\cite{icseZhangH21}.

Reducing software unfairness has become an ethical duty of software researchers and engineers \cite{fairsmotepaper,sigsoftChenZSH22}. The SE community is endeavoring to address unfairness issues in ML software \cite{fairsmotepaper,sigsoftChenZSH22}. 
In the SE domain, unfairness issues are also referred to as `fairness bugs' \cite{Dabs220710223}. SE researchers have been extensively exploring  various techniques to fix fairness bugs and improve software fairness~\cite{maxasefairness,sigsoftHortZSH21,icseZhangH21,biswas2020machine,fairwaypaper,fairsmotepaper,sigsoftBiswasR21,icseGoharBR23}. 

In practice, software systems can have multiple protected attributes that need to be considered simultaneously \cite{sigsoftChenZSH22}. From the humanities' perspective, unfair software systems built into society lead to systematic disadvantages along multiple intersecting attributes, such as sex, race, age, disability status, and so on~\cite{icdeFouldsIKP20}. 
From the SE perspective, these protected attributes pose multiple fairness requirements, some of which can be competing or conflicting, raising issues of negotiation, mediation, and conflict resolution for software engineers \cite{FinkelsteinHMRZ08}.

The intersection of these attributes creates different levels of privilege or disadvantage for various possible subgroups. For instance, black women may be vulnerable to both sexism and racism~\cite{crenshaw1989demarginalizing}. To cater for this, the literature measures intersectional fairness as the maximum disparity between subgroups that combine membership from different protected attributes \cite{GhoshGR21,sigsoftZhang022}. Intersectional fairness has been encoded in legal regulations \cite{usequal}. It clearly has implications for software researchers and engineers, who must consider the fairness regarding multiple protected attributes simultaneously as multiple non-functional software requirements. 

However, the current software fairness literature is lacking in this critical aspect. Existing fairness improvement research mostly focuses on singleton sets of protected attributes \cite{icseZhangH21,sigsoftBiswasR21,sigsoftHortZSH21,icseGoharBR23,Dabs220703277,fairwaypaper,biswas2020machine,icseLiMC0WZX22}. Unfortunately, the implications of this prevalent practice remain unclear. We have yet to fully understand the potential impact on desirable fairness properties concerning other protected attributes when catering for fairness according to a single protected attribute. Moreover, considering the legal and ethical fairness requirements~\cite{GhoshGR21,usequal,SarroRE23}, there is an urgent demand to apply fairness improvement methods to deal with multiple protected attributes. Consequently, a comprehensive study on the effectiveness of these methods in such situations becomes imperative.

Furthermore, there is an important interplay between fairness and other functional SE requirements. Specifically, it is widely recognized that fairness improvement typically comes at the cost of ML performance (e.g., accuracy), known as the fairness-performance trade-off~\cite{berk2021fairness,nipsWickpT19,CorbettDaviesP17,sigsoftHortZSH21,Dabs220703277}. Based on the current literature, it remains unclear how existing fairness improvement methods would trade-off between fairness and performance when multiple protected attributes are considered.

To fill these gaps in the literature, we conduct an extensive study of fairness improvement regarding multiple protected attributes, with 11 state-of-the-art fairness improvement methods. We evaluate these methods on five widely-adopted datasets, which cover financial, social, and medical application domains, with widely-studied ML models, fairness metrics, and performance metrics. We investigate the effect of these methods on the fairness regarding unconsidered protected attributes. We also check the performance decrease when multiple protected attributes are considered. We analyze the effectiveness of these methods for intersectional fairness improvement and fairness-performance trade-offs. If our study reveals their effectiveness, no alternative approaches would be needed; otherwise, one can build on our study's results to seek improvements in current methods or to devise novel methods that could better tackle the problem at hand.


Our study reveals the following findings: \textbf{1)} Existing methods can largely decrease fairness regarding unconsidered protected attributes. This decrease happens in up to 88.3\% of scenarios (57.5\%  on average), with a significantly large effect in up to 69.2\% of scenarios (29.1\% on average).
\textbf{2)} There is a similar decrease in accuracy when considering single and multiple protected attributes, with a 0.3\% difference in decrease rate. However, F1-score is greatly affected, with the impact on F1-score when dealing with two protected attributes being about twice that of a single attribute.
\textbf{3)} According to a state-of-the-art benchmarking tool \cite{sigsoftHortZSH21}, existing methods outperform the fairness-performance trade-off baseline constructed by the tool in 9.0\%$\sim$81.2\% of cases (52.4\% on average) when dealing with multiple protected attributes. These methods even decrease both intersectional fairness and ML performance in 6.4\%$\sim$41.6\% of cases (18.4\% on average).

Additionally, our results on the effectiveness of each studied method in improving intersectional fairness and fairness-performance trade-offs offer references for software engineers when selecting fairness improvement methods. Furthermore, these results can serve as easy-to-access baselines for researchers to evaluate new fairness improvement methods.

In summary, this paper makes the following contributions:

\begin{itemize}[leftmargin=*]
\item A rigorous empirical study on the impact of fairness improvement methods on fairness regarding unconsidered protected attributes.
\item An extensive study of the effectiveness of state-of-the-art fairness improvement methods in enhancing intersectional fairness and achieving fairness-performance trade-offs when considering multiple protected attributes.
\item A publicly-available package \cite{githublink}, containing all scripts and data in this study, to facilitate replication and extension.
\end{itemize}

\section{Preliminaries}\label{preli}
We start with introducing the background knowledge of this study.


\subsection{Protected Attributes and Fairness}\label{fairness_metric}
Fairness has emerged as an important research topic in the SE research community, with a particular focus on fairness of ML software \cite{jieMLsurvey}. The ML software fairness literature primarily concentrates on ML classification that predicts class labels for individuals based on their personal features \cite{icseZhangH21, fairsmotepaper, Dabs220703277, sigsoftChenZSH22, fairmaskpaper,icseBiswasR23}. These class labels can be categorized as favorable or unfavorable. For instance, in the context of credit scoring, a good credit is considered a favorable label, while a bad credit is deemed unfavorable.

During the classification, certain personal attributes need to be protected against discrimination. These attributes are referred to as protected attributes, also known as sensitive attributes. Common protected attributes include sex, race, age, religion, disability status, and national origin. In real-world applications, ML software often needs to consider multiple protected attributes simultaneously.

Based on the value of a protected attribute, individuals can be divided into a privileged group and an unprivileged group. In practice, the privileged group tends to be associated with favorable labels, while the unprivileged group is more likely to receive unfavorable labels. For example, in credit scoring tasks, race is often considered a protected attribute \cite{andreeva2004impact}. Due to potential biases favoring the white group in the credit scoring models, the white group may be viewed as privileged, while the non-white group may be considered unprivileged.

To address such biases, legal regulations and the fairness literature advocate for group fairness \cite{csurMehrabiMSLG21,nipsWickpT19}, which requires ML software to treat privileged and unprivileged groups equally. Mathematical metrics have been developed to measure group fairness. We describe three metrics that have been widely adopted in the software fairness literature \cite{sigsoftHortZSH21, biswas2020machine, sigsoftBiswasR21,sigsoftChenZSH22,Dabs220703277}:
\begin{itemize}[leftmargin=*]
\item \textbf{SPD} (Statistical Parity Difference) calculates the disparity in favorable rates between the privileged and unprivileged groups.
\item \textbf{AOD} (Average Odds Difference) captures the average discrepancy in false-positive rates and true-positive rates between the privileged and unprivileged groups.
\item \textbf{EOD} (Equal Opportunity Difference) assesses the disparity in true-positive rates between the privileged and unprivileged groups.
\end{itemize}

\begin{table}[!tp]
\small
\centering
\caption{Fairness metrics.}
\label{fmetric_info}
\begin{tabular}{l|c}
\toprule
Metric & Definition\\
\midrule
SPD & $P[\hat{Y} = 1 | A = 0] - P[\hat{Y} = 1 | A = 1]$\\
\midrule
AOD  & \makecell[c]{$\frac{1}{2}(P[\hat{Y}=1| A=0, Y=0]  -  P[\hat{Y}=1|A=1, Y=0]$\\ $+  P[\hat{Y}=1| A=0, Y=1]  -  P[\hat{Y}=1|A=1, Y=1])$}\\
\midrule
EOD & $P[\hat{Y}=1| A=0, Y=1]  -  P[\hat{Y}=1|A=1, Y=1]$\\
\bottomrule
\end{tabular}
\end{table}

\begin{table}[!tp]
\small
\centering
\caption{Intersectional fairness metrics.}
\label{ifmetric_info}
\begin{tabular}{l|c}
\toprule
Metric & Definition\\
\midrule
SPD & $\max \limits_{s\in S} P[\hat{Y} = 1 | A = s] - \min \limits_{s\in S} P[\hat{Y} = 1 | A = s]$\\
\midrule
AOD  & \makecell[c]{$\frac{1}{2}[\max \limits_{s\in S}(P[\hat{Y}=1| A=s, Y=0] + P[\hat{Y}=1| A=s, Y=1])$\\ $-  \min \limits_{s\in S}( P[\hat{Y}=1|A=s, Y=0]  +  P[\hat{Y}=1|A=s, Y=1])]$}\\
\midrule
EOD & $\max \limits_{s\in S} P[\hat{Y}=1| A=s, Y=1]  -  \min \limits_{s\in S} P[\hat{Y}=1|A=s, Y=1]$\\
\bottomrule
\end{tabular}
\end{table}

Let $A$ represent the protected attribute, with 1 denoting the privileged group and 0 denoting the unprivileged group. Let $Y$ denote the actual label and $\hat{Y}$ denote the predicted label, where 1 is the favorable class and 0 is the unfavorable class. The calculation methods of these fairness metrics are shown in Table \ref{fmetric_info}.








\subsection{Intersectional Fairness}
To consider multiple protected attributes and their intersectionality, researchers divide a population into subgroups based on the combination of different protected attributes \cite{GhoshGR21,sigsoftZhang022,icdeFouldsIKP20}. The intersectional fairness is measured as the maximum disparity between any two subgroups \cite{GhoshGR21,sigsoftZhang022}.
For instance, considering two protected attributes Sex = \{Male, Female\} and Race = \{White, Non-White\}, the subgroup set S = \{(Male, White), (Male, Non-White), (Female, White), (Female, Non-White)\}. If the favorable rates for the four subgroups are 50\%, 40\%, 30\%, and 20\%, SPD is calculated as $50\%-20\%=30\%$. 

Specifically, in the context of intersectional fairness, \textbf{SPD} measures the maximum difference between subgroups in obtaining favorable outcomes; \textbf{AOD} measures the maximum of the average of differences in false-positive rates and true-positive rates between subgroups; \textbf{EOD} measures the maximum difference between subgroups in true-positive rates.

Formally, we use $A$ to denote the protected attributes and define $S$ as the set of all possible combinations of the protected attributes. Let $s$  be a subgroup, where $s \in S$. These intersectional fairness metrics are calculated as shown in Table \ref{ifmetric_info}.







Compared to single-attribute fairness, intersectional fairness can capture unfairness amplified in subgroups that combine membership from different unprivileged groups \cite{GhoshGR21}, especially if such subgroups are particularly underrepresented in historical platforms of opportunity, e.g., the (Female, Non-White) subgroup in the aforementioned example.

\section{Experimental Setup}\label{experiment}
In this section, we describe our research questions and experimental settings for the study.

\subsection{Research Questions}
\noindent \textbf{RQ1:} \emph{How do existing fairness improvement methods affect the fairness regarding unconsidered protected attributes?} This RQ investigates the negative side effect of single-attribute fairness improvement by studying its impact on fairness regarding the unconsidered protected attributes.

\noindent \textbf{RQ2:} \emph{What intersectional fairness do existing fairness improvement methods achieve when considering multiple protected attributes?} This RQ evaluates the effectiveness of state-of-the-art fairness improvement methods in improving intersectional fairness.

\noindent \textbf{RQ3:} \emph{What fairness-performance trade-off do existing fairness improvement methods achieve when considering multiple protected attributes?} This RQ explores whether fairness improvement for multiple protected attributes can bring more decrease in ML performance and how state-of-the-art methods make the trade-off between intersectional fairness and ML performance. 

\noindent \textbf{RQ4:} \emph{How well do existing fairness improvement methods apply to different decision tasks, ML models, and fairness and performance metrics, when dealing with multiple protected attributes?} This RQ enriches the empirical knowledge of RQ2 and RQ3, and  explores whether existing methods are widely applicable.


\begin{table*}[!tp]
\small
\centering
\caption{Datasets.}
\label{dataset_info}
\begin{tabular}{l|rrcrl}
\toprule
Name & \#Samples & \#Features & {\makecell[c]{Protected\\attributes}} & {\makecell[r]{Favorable label \\(Proportion)}} & Task\\
\midrule
Adult  & 48,843 & 7 & sex, race & income > 50k (23.9\%) & Predicting if a person's income is greater than \$50k\\
Compas & 7,214 & 7 & sex, race & no recidivism (54.9\%) & Predicting if a criminal defendant will re-offend\\
Default & 30,000 & 23 & sex, age & default (22.1\%) & Predicting if a customer will default on payment\\
Mep15 & 15,830  & 42 & sex, race & utilizer (17.2\%) & Predicting healthcare utilization of a person\\
Mep16 & 15,675  & 42 & sex, race & utilizer (16.8\%) & Predicting healthcare utilization of a person\\
\bottomrule
\end{tabular}
\end{table*}

\subsection{Datasets and Models}\label{data_and_model}
We use five real-world datasets for study: \textbf{Adult} \cite{adultdata}, \textbf{Compas} \cite{compasdata}, \textbf{Default} \cite{defaultdata}, \textbf{Mep15} \cite{mep15data}, and \textbf{Mep16} \cite{mep16data}. A description of each dataset is presented in Table \ref{dataset_info}. These datasets have been widely adopted in the fairness literature \cite{ sigsoftZhang022,Dabs220703277,sigsoftChenZSH22,fairmaskpaper,fairsmotepaper}. They encompass tasks that involve individuals' personal information across diverse fairness-critical domains, such as finance, social, and medical. 
In line with previous fairness research \cite{fairsmotepaper, sigsoftChenZSH22, fairmaskpaper, sigsoftZhang022,Dabs220703277}, we select the two protected attributes provided by each dataset for our study.

For each dataset, we train four ML models, including Logistic Regression (LR), Support Vector Machine (SVM), Random Forest (RF), and Deep Neural Network (DNN), which have been extensively adopted in fairness literature \cite{sigsoftHortZSH21, biswas2020machine, icseZhangH21, fairsmotepaper, sigsoftChenZSH22, sigsoftZhang022, icseZhangW0D0WDD20, icseZhengCD0CJW0C22}. LR, SVM, and RF use default configurations from relevant studies \cite{sigsoftChenZSH22, icseZhangH21,sigsoftHortZSH21,Dabs220703277}, while the DNN employs a fully-connected architecture with five hidden layers, containing 64, 32, 16, 8, and 4 units, respectively, which has been widely used in previous fairness research involving similar datasets \cite{sigsoftZhang022, icseZhangW0D0WDD20, icseZhengCD0CJW0C22}.

\subsection{Fairness Improvement Methods}\label{methodused}
We employ 11 state-of-the-art fairness improvement methods for study, covering pre-processing, in-processing, and post-processing methods. 
Pre-processing methods focus on reducing bias in training data to achieve a fairer model; in-processing methods optimize training algorithms to enhance fairness; post-processing methods modify ML model predictions to ensure fair outcomes~\cite{Dabs220703277,DBcorrabs220707068}. 

First, we use eight state-of-the-art methods proposed in the ML literature~\cite{sigsoftZhang022}.

\noindent Pre-processing methods:
\begin{itemize}[leftmargin=*]
\item \textbf{RW} (Reweighting) \cite{rewpaper} employs differential weighting of training data for each combination of groups and labels to achieve fairness.
\item \textbf{DIR} (Disparate Impact Remover) \cite{kddFeldmanFMSV15} adjusts feature values to enhance fairness while preserving the rank-ordering within groups.
\end{itemize}

\noindent In-processing methods:
\begin{itemize}[leftmargin=*]
\item \textbf{META} (Meta Fair Classifier) \cite{mfcpaper} employs a meta-algorithm to optimize fairness regarding protected attributes.
\item \textbf{ADV} (Adversarial Debiasing) \cite{ADVpaper} uses adversarial techniques to minimize the presence of protected attributes in predictions, while concurrently maximizing prediction accuracy.
\item \textbf{PR} (Prejudice Remover) \cite{PRpaper} incorporates discrimination-aware regularization to mitigate the influence of protected attributes.
\end{itemize}

\noindent Post-processing methods:
\begin{itemize}[leftmargin=*]
\item \textbf{EOP} (Equalized Odds Processing) \cite{EOpaper} uses linear programming to calculate probabilities for adjusting output labels, aiming to optimize equalized odds concerning protected attributes.
\item \textbf{CEO} (Calibrated Equalized Odds) \cite{COpaper} optimizes the probabilities of modifying output labels based on calibrated classifier score outputs, with the objective of achieving equalized odds.
\item \textbf{ROC} (Reject Option Classification) \cite{ROCpaper} assigns favorable outcomes to unprivileged instances and unfavorable outcomes to privileged instances near the decision boundary, particularly when there is high uncertainty.
\end{itemize}

Second, we use three state-of-the-art methods proposed in the SE literature, including Fair-SMOTE~\cite{fairsmotepaper}, MAAT~\cite{sigsoftChenZSH22}, and FairMask~\cite{fairmaskpaper}. 

\begin{itemize}[leftmargin=*]
\item \textbf{Fair-SMOTE} \cite{fairsmotepaper} generates synthetic samples to achieve balanced distributions not only between different labels but also among various protected attributes within the training data. Additionally, it removes ambiguous samples from the training set.
\item \textbf{MAAT} \cite{sigsoftChenZSH22} combines individual models optimized for ML performance and fairness concerning each protected attribute, respectively. It ensures that both fairness and ML performance objectives are met.
\item \textbf{FairMask} \cite{fairmaskpaper} trains extrapolation models to predict protected attributes based on other data features. Subsequently, it uses these extrapolation models to modify the protected attributes in test data, enabling fairer predictions.
\end{itemize}

We apply each fairness improvement method to the original models obtained in Section \ref{data_and_model}. 
We repeat each experiment 20 times. Each time we randomize the dataset by shuffling it and then divide it into 70\% training data and 30\% test data.

When conducting fairness improvement for multiple protected attributes, we simultaneously consider these attributes instead of applying a fairness improvement method independently for each attribute. It is because individually applying the method for each protected attribute cannot maintain fairness for previous considered attributes while also guaranteeing fairness for subsequently considered attributes. For example, let us consider a dataset with two protected attributes, and only one attribute is considered for fairness improvement at a time. Pre-processing methods may not preserve the optimized characteristics for the first considered protected attribute when optimizing data characteristics for the second attribute. For instance, if we use the RW method to assign different weights based on the second attribute, it can undermine its intended weights for the first attribute. In the case of in-processing methods, training models for one protected attribute results in models specific to that attribute. Therefore, in-processing methods can disregard fairness considerations for the first attribute, when optimizing for the second attribute. Additionally, concerning post-processing methods, modifying the output to optimize fairness for the second protected attribute may not ensure the preservation of fairness for the first attribute.

\subsection{Measurement Metrics}
We employ three fairness metrics and five ML performance metrics, resulting in a total of 15 fairness-performance measurements for study, as detailed in the following.

\subsubsection{Fairness metrics}
We use three fairness metrics introduced in Section \ref{preli}, including \textbf{SPD}, \textbf{AOD}, and \textbf{EOD}, which have been widely adopted in the fairness literature \cite{sigsoftHortZSH21, biswas2020machine, sigsoftBiswasR21,sigsoftChenZSH22,Dabs220703277}. We calculate the fairness metric values for individual attributes and intersectional fairness, as listed in Tables~\ref{fmetric_info} and \ref{ifmetric_info}.
We use absolute values for all fairness metrics, whereby these metrics indicate the highest fairness when they equal 0, and larger values indicate greater unfairness.

\subsubsection{Performance metrics}
We follow previous work \cite{sigsoftChenZSH22,Dabs220703277} to use a comprehensive set of five common ML performance metrics for study: \textbf{accuracy}, \textbf{precision}, \textbf{recall}, \textbf{F1-score}, and \textbf{MCC} (Matthews Correlation Coefficient). We provide the formal definitions of these metrics in Table \ref{pmetric_info}, where TP, TN, FP, and FN denote the number of true positives, true negatives, false positives, and false negatives, respectively. For precision, recall, and F1-score, we report the macro-average values, as done in previous research \cite{sigsoftChenZSH22}, to enable comparisons of overall performance on the favorable and unfavorable classes. To achieve this, we average the precision, recall, and F1-score results obtained for the two classes.  For each of the five metrics, a higher value indicates better ML performance. 


\begin{table}[!tp]
\scriptsize
\centering
\caption{ML performance metrics.}
\label{pmetric_info}
\begin{tabular}{l|c}
\toprule
Metric & Definition\\
\midrule
Accuracy & $(TP+TN)/(TP+FP+TN+FN)$\\
Precision  & $TP/(TP+FP)$\\
Recall & $TP/(TP+FN)$\\
F1-score & $2\times Precision \times Recall/(Precision + Recall)$\\
MCC & $(TP \times TN-FP \times FN)/\sqrt{(TP+FP)(TP+FN)(TN+FP)(TN+FN)}$\\
\bottomrule
\end{tabular}
\end{table}

\subsubsection{Fairness-performance trade-off measurement}\label{fairea_des}
To assess the fairness-performance trade-off, we rely on \textbf{Fairea} \cite{sigsoftHortZSH21}, a state-of-the-art benchmarking tool that offers a unified trade-off baseline for comparing various fairness improvement methods.

Fairea visualizes fairness and performance values using a two-dimensional coordinate system and establishes the trade-off baseline by connecting fairness-performance points of the original ML model and a set of mutated models. The mutated models are generated by gradually transforming the original model into models that produce only the majority class in the dataset. Throughout this process, fairness improves as the predictive performance becomes equally worse for privileged and unprivileged groups. Fairea uses these naive mutated models to establish the trade-off baseline, as it expects that fairness improvement methods should outperform them.

Fairea classifies the trade-off effectiveness of fairness improvement methods into five levels by comparing the fairness-performance trade-off achieved by these methods with the established baseline:
\begin{itemize}[leftmargin=*]
\item The \emph{win-win trade-off} level includes methods that increase both fairness and performance.
\item The \emph{good trade-off} level includes methods that increase fairness, decrease performance, and achieve a better trade-off than the baseline generated by Fairea.
\item The \emph{poor trade-off} level includes methods that increase fairness, decrease performance, and achieve a worse trade-off than the baseline.
\item The \emph{inverted trade-off} level includes methods that decrease fairness but increase performance.
\item The \emph{lose-lose trade-off} level includes methods that decrease both fairness and performance.
\end{itemize}

Different from the original paper of Fairea \cite{sigsoftHortZSH21} that focuses on single-attribute tasks, our study extends the scope to multi-attribute tasks.
We conduct a comprehensive evaluation by considering 15 fairness-performance measurements (i.e., the combination of three fairness metrics and five ML performance metrics). 

For each combination of \emph{(dataset, ML model, fairness-performance measurement)}, we establish a trade-off baseline. To achieve this, we first train the original model and then generate the mutated models based on it. The process is repeated 20 times. Following the recommendation of Fairea \cite{sigsoftHortZSH21}, we determine the baseline by averaging the results of these multiple runs.

\subsection{Statistical Analysis}
We use three statistical analysis methods in this study: Mann Whitney U-test \cite{mann1947test}, Cliff's $\delta$ \cite{eseKitchenhamMBKBC17}, and Spearman’s correlation coefficient $\rho$ \cite{myers2013research}. Since these methods do not assume normality of the data, they are suitable for our study, where we deal with diverse data that may not follow a normal distribution.

In RQ1 and RQ2, we use the Mann Whitney U-test \cite{mann1947test} to assess whether fairness improvement methods significantly impact fairness. To establish statistical significance, we follow previous work~\cite{Dabs220703277,sigsoftChenZSH22} to consider a $p$-value lower than 0.05. Specifically, when comparing two sets of fairness values using the test, we conclude that the two sets have statistically different fairness if the $p$-value of the test is lower than 0.05.
Furthermore, to measure the effect size of the impact, we adopt the Cliff's $\delta$ \cite{eseKitchenhamMBKBC17}, a commonly-used metric in the SE literature \cite{eseKitchenhamMBKBC17, vargha2000critique, esemBenninKMPM17}. Consistent with the literature \cite{eseKitchenhamMBKBC17, vargha2000critique, esemBenninKMPM17}, we consider a change with an absolute value of $\delta$ greater than or equal to 0.428 as indicative of a large effect.
Additionally, in RQ1, we use the Spearman's correlation coefficient $\rho$ \cite{myers2013research} to explore potential factors correlated with the impact on unconsidered protected attributes. The coefficient $\rho$ ranges from -1 to 1, with 1 representing a perfect positive correlation, 0 indicating no correlation, and -1 representing a perfect negative correlation.  A correlation is considered statistically significant only when the coefficient yields a $p$-value lower than 0.05 \cite{Dabs220703277}.

\section{Results}
This section answers our RQs based on the experimental results. Due to the page limit, we primarily report statistical results in the paper and include the results of each fairness improvement method for each scenario in our repository~\cite{githublink}.

\subsection{RQ1: Impact on Unconsidered Protected Attributes}
This RQ investigates how fairness improvement methods affect the fairness regarding unconsidered protected attributes when targeting a single protected attribute. Each dataset-protected attribute pair, as shown in Table \ref{dataset_info}, represents a single-attribute fairness improvement task. For instance, in the case of the Adult dataset, we have two tasks: Adult-Sex and Adult-Race. We apply existing methods to improve fairness for one task and then examine the influence on the fairness of the other task. 
Each application is repeated 20 times using four ML models and three fairness metrics (more details in Section~\ref{experiment}). We treat each combination of \emph{(task, ML model, fairness metric)} as a scenario and calculate the proportions of scenarios where existing methods reduce fairness regarding unconsidered protected attributes, based on the average value obtained from the 20 repeated runs.


Table \ref{fp_alone} shows the results. 
The methods that we study decrease fairness regarding unconsidered protected attributes in up to 88.3\% of the total scenarios (with an average of 57.5\% across different methods).
We further analyze the significance and effect size of the decrease by using Mann Whitney U-test and Cliff's $\delta$, and find that such decrease has a significantly large effect in up to 69.2\% of the scenarios (29.1\% on average). 

We take the three methods highlighted in Table \ref{fp_alone} as examples to illustrate why they cause a large fairness decrease for unconsidered protected attributes.
Fair-SMOTE aims to balance data for one protected attribute, which can lead to more severe data imbalance for other protected attributes, resulting in reduced fairness for those attributes. 
ROC targets predictions with high uncertainty and tends to assign favorable outcomes to the unprivileged members and unfavorable outcomes to the privileged. For example, if sex is the considered protected attribute and race is the unconsidered one, predictions for (Male, Non-White) and (Female, White) members tend to be uncertain because the two subgroups have both privileged and unprivileged properties. Therefore, improving fairness for sex can lead to more unfavorable outcomes for (Male, Non-White) and more favorable outcomes for (Female, White), causing further unfairness regarding race. META aims to improve fairness for the protected attribute during training, but this objective may conflict with fairness for other unconsidered protected attributes, resulting in reduced fairness for these attributes.


\begin{table}[!tp]
\small
\centering
\caption{(RQ1) Proportions of scenarios where existing methods reduce fairness regarding unconsidered protected attributes (the second column) and also have a significantly large effect (the third column). Significantly large reductions are highlighted in bold. The top three values in each column are shaded.  The results indicate that existing methods decrease fairness regarding unconsidered protected attributes in up to 88.3\% of scenarios (57.5\% on average across different methods), with a significantly large effect observed in up to 69.2\% of scenarios (29.1\% on average).}
\label{fp_alone}
\begin{tabular}{l|rr}
\toprule
 Method & \makecell[r]{$\downarrow$ unconsidered\\fairness} & \makecell[r]{Significantly\\large effect}\\
 \midrule
RW & 49.2\% & \textbf{2.5\%}  \\
DIR & 54.2\%& \textbf{1.7\%}  \\
META & \cellcolor{gray!50}81.7\% &\cellcolor{gray!50}\textbf{69.2\%}  \\
ADV  & 64.2\%& \textbf{40.0\%}\\
PR  & 70.8\% &\textbf{45.0\%}\\
EOP & 22.5\% &\textbf{3.3\%} \\
CEO  & 37.5\% &\textbf{4.2\%} \\
ROC  & \cellcolor{gray!50}85.0\% &\cellcolor{gray!50}\textbf{68.3\%} \\
Fair-SMOTE  & \cellcolor{gray!50}88.3\%& \cellcolor{gray!50}\textbf{63.3\%}\\
MAAT  & 29.2\% &\textbf{4.2\%} \\
FairMask  & 50.0\% &\textbf{18.3\%}\\
\midrule
Average & 57.5\% &\textbf{29.1\%} \\
\bottomrule
\end{tabular}
\end{table}


To gain further insight into the fairness reduction, we explore its potential reasons from the perspective of datasets. If the protected attributes in a dataset consistently have the same values (i.e., perfectly positively correlated), improving fairness for one protected attribute would be equivalent to doing so for the others. Drawing inspiration from this observation, we hypothesize that as the correlation between the considered and unconsidered protected attributes becomes more positive, fairness improvement methods will have a lesser adverse impact on the fairness concerning the unconsidered attributes.

To test this hypothesis, we assign 1 to denote the privileged group and 0 for the unprivileged group for each protected attribute. For each task, we calculate Spearman's correlation coefficient $\rho$ to quantify the correlation between the considered and unconsidered protected attributes. Additionally, we determine the proportions of scenarios where existing methods reduce the fairness regarding the unconsidered attributes. Then, we measure the correlation between these proportions and the correlation of protected attributes.

Table \ref{root_rq1} presents the results. Based on the results, we confirm our hypothesis with a correlation coefficient of $\rho=-0.640$ at a significance level of 0.05 ($p$-value < 0.05). We illustrate this negative correlation further using the Default dataset as an example. This dataset exhibits the most negative correlation ($\rho=-0.069$, $p$-value < 0.05) among all the datasets. Meanwhile, on this dataset, existing methods reduce fairness regarding unconsidered protected attributes in the highest proportion of scenarios.

\begin{table}[!tp]
\small
\centering
\caption{(RQ1) Correlation between considered and unconsidered protected attributes (second column) and proportions of scenarios where existing methods reduce fairness for unconsidered protected attributes (third column). * indicates a significant correlation with $p$-value < 0.05. We find that the more positive the correlation between the considered and unconsidered protected attributes, the less existing methods reduce fairness regarding the unconsidered protected attributes.}
\label{root_rq1}
\begin{tabular}{l|r|r}
\toprule
Task &\makecell[c]{Correlation between\\protected attributes} & \makecell[c]{$\downarrow$ unconsidered\\fairness} \\
\midrule
Adult-Sex & 0.101* & 44.7\% \\
Adult-Race & 0.101* & 56.8\%\\
Compas-Sex & 0.068* & 40.9\%\\
Compas-Race & 0.068* & 33.3\%\\
Default-Sex & -0.069* & 84.1\%\\
Default-Age & -0.069*  & 76.5\%\\
Mep15-Sex & -0.015 & 74.2\%\\
Mep15-Race & -0.015 & 64.4\%\\
Mep16-Sex & -0.016* & 48.5\%\\
Mep16-Race & -0.016* & 51.5\%\\
\midrule 
\multicolumn{2}{c}{Correlation between the last two columns} & -0.640*\\
\bottomrule
\end{tabular}
\end{table}

\finding{The fairness improvement methods that we study can lead to decreased fairness regarding unconsidered protected attributes to a large extent. Specifically, the decrease occurs in up to 88.3\% of scenarios (on average 57.5\%), with a significantly large effect in up to 69.2\% of scenarios (on average 29.1\%). Our correlation analysis suggests that the more positive the correlation between the considered and unconsidered protected attributes, the less existing methods reduce fairness regarding the unconsidered protected attributes.}

\subsection{RQ2: Intersectional Fairness Improvement}
This RQ aims to evaluate the effectiveness of existing methods in improving intersectional fairness when dealing with multiple protected attributes. To this end, we use five datasets from Table~\ref{dataset_info}, along with four ML models and three fairness metrics for each dataset. 
We consider each \emph{(dataset, model, fairness metric)} combination as a scenario and calculate the proportions of scenarios where existing methods improve intersectional fairness based on the average results of 20 repeated runs. We also report the proportions of scenarios where the improvement has a significantly large effect by using Mann Whitney U-test and Cliff's $\delta$.

\begin{table}[!tp]
\small
\centering
\caption{(RQ2) Proportions of scenarios where existing methods improve intersectional fairness (the second column) and also have a significantly large effect (the third column). The proportions of scenarios where such improvement has a significantly large effect are highlighted in bold. The top three values in each column are shaded. MAAT, FairMask, and RW improve intersectional fairness in the most scenarios.}
\label{fairness_increase}
\begin{tabular}{l|rr}
\toprule
Method & \makecell[r]{$\uparrow$ intersectional\\fairness}  & \makecell[r]{Significantly\\large effect}\\
 \midrule
RW & \cellcolor{gray!50}90.0\%  & \cellcolor{gray!50}\textbf{68.3\%} \\
DIR & 68.3\% & \textbf{43.3\%}\\
META & 28.3\% & \textbf{16.7\%}\\
ADV & 36.7\% & \textbf{31.7\%}\\
PR & 75.0\% & \textbf{53.3\%}\\
EOP & 85.0\% & \textbf{51.7\%} \\
CEO & 61.7\% & \textbf{43.3\%} \\
ROC  & 6.7\% & \textbf{0.0\%}  \\
Fair-SMOTE & 56.7\% & \textbf{38.3\%}\\
MAAT & \cellcolor{gray!50}98.3\% & \cellcolor{gray!50}\textbf{71.7\%} \\
FairMask & \cellcolor{gray!50}93.3\% & \cellcolor{gray!50}\textbf{68.3\%}\\
\bottomrule
\end{tabular}
\end{table}

Table \ref{fairness_increase} presents the results. The 11 methods studied improve intersectional fairness in a wide range of scenarios, ranging from 6.7\% to 98.3\%. In particular, MAAT, FairMask, and RW exhibit the most consistent improvements, achieving this in 98.3\%, 93.3\%, and 90.0\% of scenarios, respectively. Furthermore,  these three methods significantly improve intersectional fairness with a large effect in the most scenarios, accounting for 71.7\%, 68.3\%, and 68.3\%, respectively. The superiority of these methods can be attributed to their ability to mitigate data bias, preventing its amplification during training or decision-making. However, a common limitation of them is the need for access of training data.  In situations where obtaining such access is infeasible, (e.g., due to privacy concerns), practitioners may prefer using post-processing methods that modify prediction outcomes to ensure fairness without requiring access to training data. Among the post-processing methods studied, EOP stands out, improving intersectional fairness in the most scenarios (85.0\%), with a significantly large effect in 51.7\% of cases.

We further measure the effectiveness of existing methods by calculating the absolute and relative changes in fairness metric values. Table \ref{fairness_change} presents the results averaged over the five datasets and four models under study. Methods that lower fairness metric values to the largest extent contribute the most to improving intersectional fairness, as smaller fairness metric values indicate reduced unfairness. Notably, MAAT and FairMask, two state-of-the-art methods from the SE literature, demonstrate a general advantage in enhancing intersectional fairness across various fairness metrics. Specifically, they improve AOD fairness by 32.4\% and 34.9\%, respectively. Additionally, RW, PR, and EOP also yield favorable results in specific fairness metrics. Among the highlighted methods, EOP, as a post-processing method, is the only one that does not require access to training data. This makes EOP a suitable choice for scenarios where obtaining such access is infeasible.

\begin{table}[!tp]
\footnotesize
\centering
\caption{(RQ2) Absolute and relative changes (in parentheses) in intersectional fairness achieved by existing methods. The top three values in each column are highlighted.  MAAT and FairMask demonstrate superiority in improving intersectional fairness across different fairness metrics.}
\label{fairness_change}
\begin{tabular}{l|rr|rr|rr}
\toprule
Method & \multicolumn{2}{c|}{SPD} & \multicolumn{2}{c|}{AOD} & \multicolumn{2}{c}{EOD}\\
 \midrule
RW & -0.030 &(-20.0\%) & \cellcolor{gray!50}-0.036 &\cellcolor{gray!50}(-25.7\%) & -0.040& \cellcolor{gray!50}(-22.8\%) \\
DIR & -0.026 &(-14.6\%) & -0.024 &(-7.6\%) & -0.032 &(-3.0\%) \\
META & 0.192 & (236.4\%) & 0.139& (202.4\%) & 0.048 &(66.0\%) \\
ADV & -0.004& (-3.7\%) & 0.032& (48.6\%) & 0.046& (57.5\%)\\
PR & \cellcolor{gray!50}-0.056& \cellcolor{gray!50}(-41.7\%) & -0.031& (-13.1\%) & \cellcolor{gray!50}-0.046& (-6.9\%) \\
EOP & \cellcolor{gray!50}-0.038& (-20.6\%) & -0.035& (-20.5\%) & -0.037& (-17.2\%) \\
CEO & -0.006& (-9.5\%) & -0.003& (-8.4\%) & -0.022& (-10.0\%) \\
ROC & 0.104& (80.2\%) & 0.085& (64.6\%) & 0.069& (39.8\%) \\
Fair-SMOTE & -0.004& (15.2\%) & -0.028& (-4.7\%) & -0.038& (-14.9\%) \\
MAAT & \cellcolor{gray!50}-0.041& \cellcolor{gray!50}(-29.3\%) & \cellcolor{gray!50}-0.042& \cellcolor{gray!50}(-32.4\%) & \cellcolor{gray!50}-0.054 &\cellcolor{gray!50}(-29.2\%) \\
FairMask & -0.034& \cellcolor{gray!50}(-22.2\%) & \cellcolor{gray!50}-0.050& \cellcolor{gray!50}(-34.9\%) & \cellcolor{gray!50}-0.067 &\cellcolor{gray!50}(-32.0\%) \\
\bottomrule
\end{tabular}
\end{table}

\finding{The fairness improvement methods that we study improve intersectional fairness in 6.7\%$\sim$98.3\% of the scenarios. Notably, MAAT, FairMask, and RW achieve this goal in the most scenarios, accounting for 98.3\%, 93.3\%, and 90.0\%, respectively; the improvement has a significantly large effect in 71.7\%, 68.3\%, and 68.3\% of scenarios. For applications where obtaining access to training data is impossible (e.g., due to privacy concerns), EOP can be a better option, which improves intersectional fairness in 85.0\% of scenarios, with a significantly large effect in 51.7\% of scenarios.}

\subsection{RQ3: Fairness-performance Trade-off}
This RQ aims to evaluate the fairness-performance trade-off achieved by existing methods when dealing with multiple protected attributes. We investigate this RQ by answering two sub-questions.

\begin{table*}[!tp]
\scriptsize
\centering
\caption{(RQ3.1) Absolute and relative changes (in parentheses) in ML performance when existing methods improve fairness for single or multiple protected attributes. On average, the accuracy decrease is similar when considering single or multiple protected attributes, with only a 0.3\% difference in decrease rate. However, F1-score and MCC show significant variations between the two scenarios.}
\label{per_decrease}
\begin{tabular}{l|rr|rr|rr|rr|rr}
\toprule
\multirow{2}{*}{Method} & \multicolumn{2}{c|}{Accuracy} & \multicolumn{2}{c|}{Precision} & \multicolumn{2}{c|}{Recall} & \multicolumn{2}{c|}{F1-score} & \multicolumn{2}{c}{MCC}\\
 & single-attr & multi-attr  & single-attr & multi-attr & single-attr & multi-attr & single-attr & multi-attr & single-attr & multi-attr\\
\midrule
RW & -0.001 (-0.2\%) & -0.001 (-0.2\%) & 0.001 (0.2\%) & -0.001 (-0.1\%) & -0.003 (-0.4\%) & 0.002 (0.3\%) & -0.002 (-0.3\%) & 0.002 (0.2\%) & -0.003 (-0.5\%) & 0.001 (0.2\%)\\
DIR & -0.004 (-0.5\%) & -0.008 (-0.9\%) & -0.002 (-0.3\%) & -0.008 (-1.0\%) & -0.010 (-1.5\%) & -0.019 (-2.8\%) & -0.013 (-2.0\%) & -0.024 (-3.6\%) & -0.018 (-4.8\%) & -0.036 (-9.5\%)\\
META & -0.063 (-7.4\%) & -0.079 (-9.4\%) & -0.064 (-8.3\%) & -0.080 (-10.3\%) & 0.056 (8.8\%) & 0.050 (7.9\%) & 0.000 (0.3\%) & -0.011 (-1.4\%) & 0.001 (1.8\%) & -0.020 (-3.8\%)\\
ADV & 0.002 (0.3\%) & 0.000 (0.0\%) & 0.005 (0.7\%) & 0.004 (0.6\%) & 0.005 (0.9\%) & -0.002 (-0.1\%) & 0.007 (1.3\%) & 0.000 (0.3\%) & 0.010 (3.9\%) & 0.001 (1.8\%)\\
PR & -0.003 (-0.4\%) & -0.014 (-1.6\%) & 0.012 (1.7\%) & 0.033 (4.3\%) & -0.021 (-3.2\%) & -0.058 (-8.5\%) & -0.023 (-3.4\%) & -0.074 (-10.4\%) & -0.023 (-5.2\%) & -0.085 (-18.5\%)\\
EOP & -0.013 (-1.7\%) & -0.012 (-1.6\%) & -0.018 (-2.4\%) & -0.016 (-2.2\%) & -0.018 (-2.7\%) & -0.014 (-2.0\%) & -0.020 (-2.8\%) & -0.015 (-2.1\%) & -0.037 (-9.1\%) & -0.030 (-7.1\%)\\
CEO & -0.010 (-1.3\%) & -0.005 (-0.6\%) & -0.008 (-1.0\%) & -0.008 (-1.0\%) & -0.027 (-4.0\%) & -0.014 (-2.0\%) & -0.033 (-4.6\%) & -0.016 (-2.2\%) & -0.045 (-10.5\%) & -0.023 (-5.1\%)\\
ROC & -0.056 (-6.7\%) & -0.051 (-6.1\%) & -0.069 (-9.0\%) & -0.065 (-8.6\%) & 0.060 (9.4\%) & 0.057 (8.9\%) & 0.007 (1.3\%) & 0.010 (1.7\%) & 0.004 (1.9\%) & 0.004 (2.0\%)\\
Fair-SMOTE & -0.044 (-5.4\%) & -0.044 (-5.4\%) & -0.061 (-7.9\%) & -0.060 (-7.9\%) & 0.051 (7.9\%) & 0.046 (7.2\%) & 0.011 (1.8\%) & 0.009 (1.5\%) & 0.001 (0.5\%) & -0.004 (-0.5\%)\\
MAAT & 0.000 (-0.1\%) & -0.003 (-0.4\%) & 0.009 (1.2\%) & 0.006 (0.7\%) & -0.009 (-1.3\%) & -0.020 (-3.0\%) & -0.007 (-1.0\%) & -0.021 (-2.9\%) & -0.005 (-0.9\%) & -0.023 (-5.4\%)\\
FairMask & -0.001 (-0.2\%) & -0.003 (-0.4\%) & 0.000 (0.0\%) & -0.004 (-0.5\%) & 0.001 (0.1\%) & -0.001 (-0.2\%) & 0.001 (0.2\%) & -0.002 (-0.2\%) & 0.000 (0.4\%) & -0.005 (-0.9\%)\\
\midrule
Average & -0.018 (-2.1\%) & -0.020 (-2.4\%) & -0.018 (-2.3\%) & -0.018 (-2.4\%) & 0.008 (1.3\%) & 0.002 (0.5\%) & -0.007 (-0.8\%) & -0.013 (-1.7\%) & -0.011 (-2.0\%) & -0.020 (-4.3\%)\\
\bottomrule
\end{tabular}
\end{table*}



\subsubsection{RQ3.1: Does the application of existing methods to improve fairness for multiple protected attributes lead to significantly greater performance reduction compared to improving fairness for a single attribute? } It is well known that fairness improvement often comes at the expense of ML performance \cite{berk2021fairness,nipsWickpT19,CorbettDaviesP17,sigsoftHortZSH21,Dabs220703277}. Intuitively, improving fairness for multiple protected attributes might result in a more substantial performance decrease than doing so for a single attribute. To explore this, we calculate the absolute and relative changes in the five performance metrics that we analyze when employing existing fairness improvement methods for one or multiple protected attributes. These changes are then averaged over the five datasets and four models used in our study.

Table \ref{per_decrease} presents the results. Different from intuition, we observe a similar accuracy decrease when considering single and multiple protected attributes. When considering two protected attributes, accuracy is further decreased by 0.3\% (from -2.1\% to -2.4\%) with an absolute change of 0.002 (from -0.018 to -0.020), compared to considering a single protected attribute. Similarly, the overall decrease in precision is also minor, with a 0.1\% difference in relative change. This indicates that accuracy and precision can be reasonably maintained in the multiple-attribute paradigm.

In contrast, both F1-score and MCC experience significant decreases. Specifically, the decrease in F1-score is twice as much (from -0.8\% to -1.7\%) when considering two protected attributes. A similar pattern of decrease is observed for MCC.

Among all five metrics, recall is the only one that shows an overall improvement when conducting fairness improvement. However, when dealing with two protected attributes, the improvement in recall decreases compared to that when dealing with a single attribute (from 1.3\% to 0.5\%).

The findings regarding performance decreases carry significant implications for the use of performance metrics in fairness research. As mentioned by previous studies \cite{Dabs220703277}, the majority of existing fairness research \cite{icseZhangH21,sigsoftHortZSH21,icdmCaldersKP09,datamineCaldersV10,mfcpaper,kddFeldmanFMSV15,rewpaper,icdmKamiranCP10,isciKamiranMKZ18,aistatsZafarVGG17} relies solely on accuracy as the performance metric. Our results demonstrate that by exclusively focusing on accuracy, researchers may overlook the significant impact on other performance metrics when dealing with multiple protected attributes. In real-world applications, metrics such as F1-score and MCC are important \cite{Dabs220703277,isstaMoussaS22}. Therefore, solely relying on accuracy may not provide engineers with a complete picture when selecting fairness improvement methods for such applications. 

Furthermore, we find that different methods can exhibit distinct performance decrease patterns. For instance, we examine the accuracy decrease of two top-performing methods identified in RQ2 (i.e., FairMask and RW) when considering two protected attributes.
FairMask, which improves fairness by modifying protected attribute information, experiences a doubled accuracy decrease (-0.2\% vs. -0.4\%) when dealing with two protected attributes. This is because FairMask needs to obfuscate more information to achieve fairness, resulting in a higher accuracy sacrifice. Compared to FairMask, RW adjusts only the weights of samples in training without modifying any attributes, avoiding introducing significant noise when dealing with more protected attributes. This characteristic enables RW to maintain a comparable accuracy when dealing with one or two protected attributes (-0.2\% vs. -0.2\%).

\finding{Different from intuition, we observe a similar accuracy decrease when considering single and multiple protected attributes (with a 0.3\% difference in decrease rate), suggesting that accuracy can be maintained in the multiple-attribute paradigm. However, F1-score and MCC are greatly affected, showing an impact about twice as great when dealing with two protected attributes compared to a single attribute. Therefore, considering only change in accuracy (as most fairness studies do) cannot provide implications for real-world applications where F1-score or MCC is crucial.}

\subsubsection{RQ3.2: Which trade-off effectiveness levels do existing fairness improvement methods fall into according to Fairea?}
In this RQ, we use Fairea \cite{sigsoftHortZSH21}, a state-of-the-art benchmarking tool described in Section \ref{fairea_des}, to evaluate the effectiveness of existing methods in achieving the trade-off between intersectional fairness and ML performance when dealing with multiple protected attributes. For each of the five datasets, we use four ML models and 15 fairness-performance measurements. We apply each fairness improvement method to the $5\times4\times15=300$ \emph{(dataset, model, measurement)} combinations. We repeat the experiments 20 times and treat each single run as an individual case. As a result, we have $300\times20=6,000$ cases for each method. We use Fairea to classify the trade-offs achieved by each method in these cases into different effectiveness levels, and then calculate the distribution of the effectiveness levels.

\begin{figure} 
    \centering
\includegraphics[width=1\linewidth]{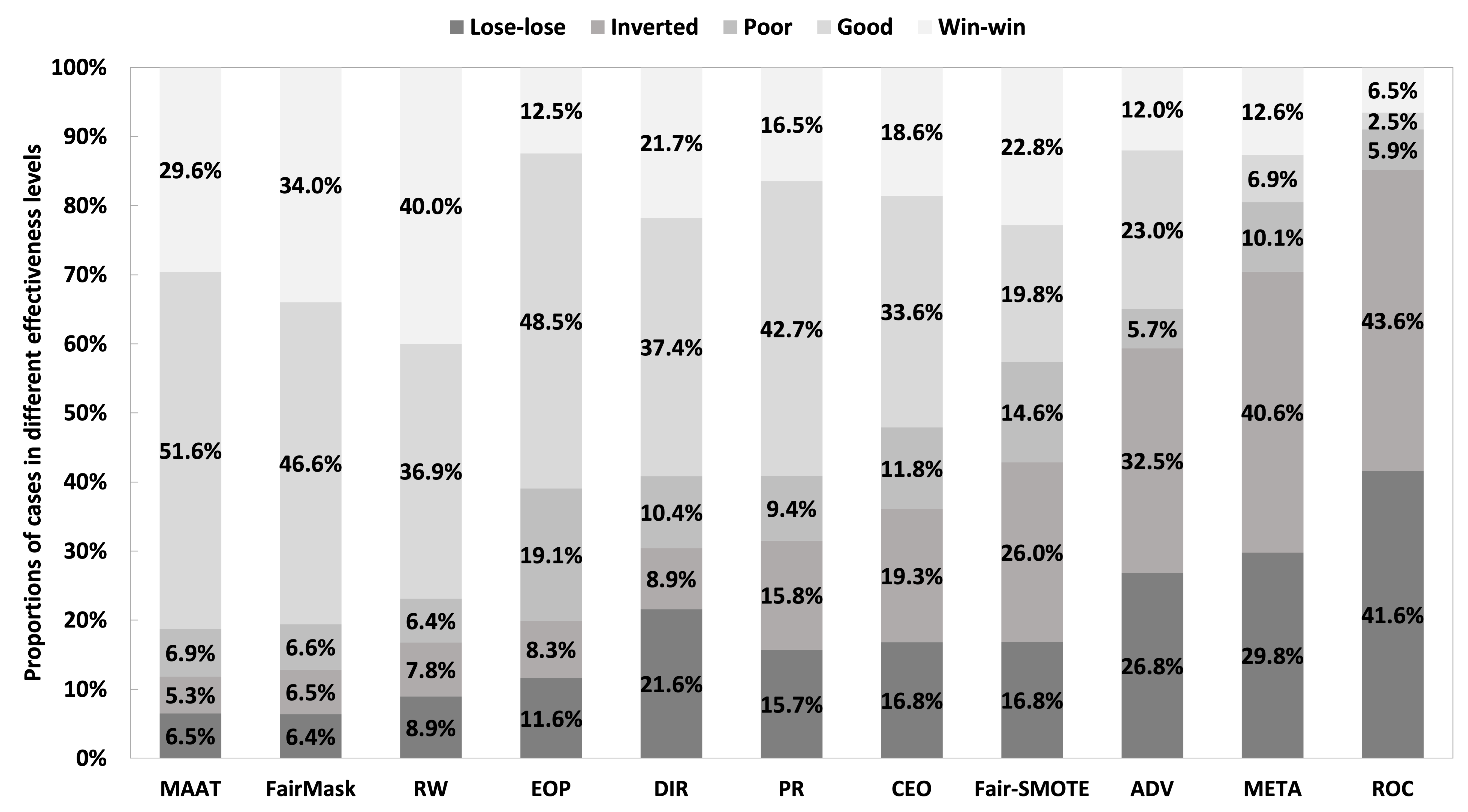}
  \caption{(RQ3.2) Effectiveness level distributions of existing methods in fairness-performance trade-off when dealing with multiple protected attributes. MAAT, FairMask, and RW achieve the best trade-off, with 81.2\%, 80.6\%, and 76.9\% of cases falling into the win-win or good trade-off, respectively.}
  \label{fig:regiondis} 
\end{figure}

We illustrate the results in Figure \ref{fig:regiondis} and present the methods in descending order by the proportion of cases where each method beats the trade-off baseline constructed by Fairea (i.e., achieving win-win or good trade-off). These methods surpass the trade-off baseline in 9.0\%$\sim$81.2\% of cases (52.4\% on average). They also achieve a lose-lose trade-off (i.e., decrease both intersectional fairness and performance) in 6.4\%$\sim$41.6\% of cases (18.4\% on average). 

Among the 11 methods under study, MAAT, FairMask, and RW achieve the best trade-off effectiveness. They beat the trade-off baseline constructed by Fairea in 81.2\%, 80.6\%, and 76.9\% of the evaluated cases, respectively. In particular, they improve both intersectional fairness and performance (i.e., win-win trade-off) in 29.6\%, 34.0\%, and 40.0\% of cases. Nevertheless, they still suffer from a lose-lose trade-off (i.e., decreasing both intersectional fairness and ML performance) in 6.5\%, 6.4\%, and 8.9\% of cases.


\finding{The state-of-the-art fairness improvement methods that we study beat the fairness-performance trade-off baseline constructed by Fairea in 9.0\%$\sim$81.2\% of cases (52.4\% on average) when dealing with multiple protected attributes. They also lead to a decrease of both intersectional fairness and performance in 6.4\%$\sim$41.6\% of cases (18.4\% on average). Among these methods, MAAT, FairMask, and RW are the most effective, surpassing the trade-off baseline in 81.2\%, 80.6\%, and 76.9\% of the evaluated cases, respectively. }


\subsection{RQ4: Applicability}
This RQ aims to explore whether existing fairness improvement methods are widely applicable to different datasets, models, and fairness-performance measurements.
Specifically, we analyze the effectiveness of these methods in improving intersectional fairness and achieving the trade-off between intersectional fairness and performance. For the effectiveness in fairness improvement, we calculate the proportions of scenarios where existing methods improve intersectional fairness for each dataset, model, and fairness measurement, respectively. For example, for each dataset, we have $4\times3=12$ \emph{(model, fairness metric)} combinations, and compute the proportion of the 12 scenarios in which each method improves intersectional fairness. For the effectiveness in the fairness-performance trade-off, we use the proportion of cases that surpass the trade-off baseline constructed by Fairea as the indicator \cite{sigsoftChenZSH22}, and calculate the proportions achieved by each method for each dataset, model, and fairness-performance measurement.  

\begin{figure*} 
    \centering
\includegraphics[width=0.65\linewidth]{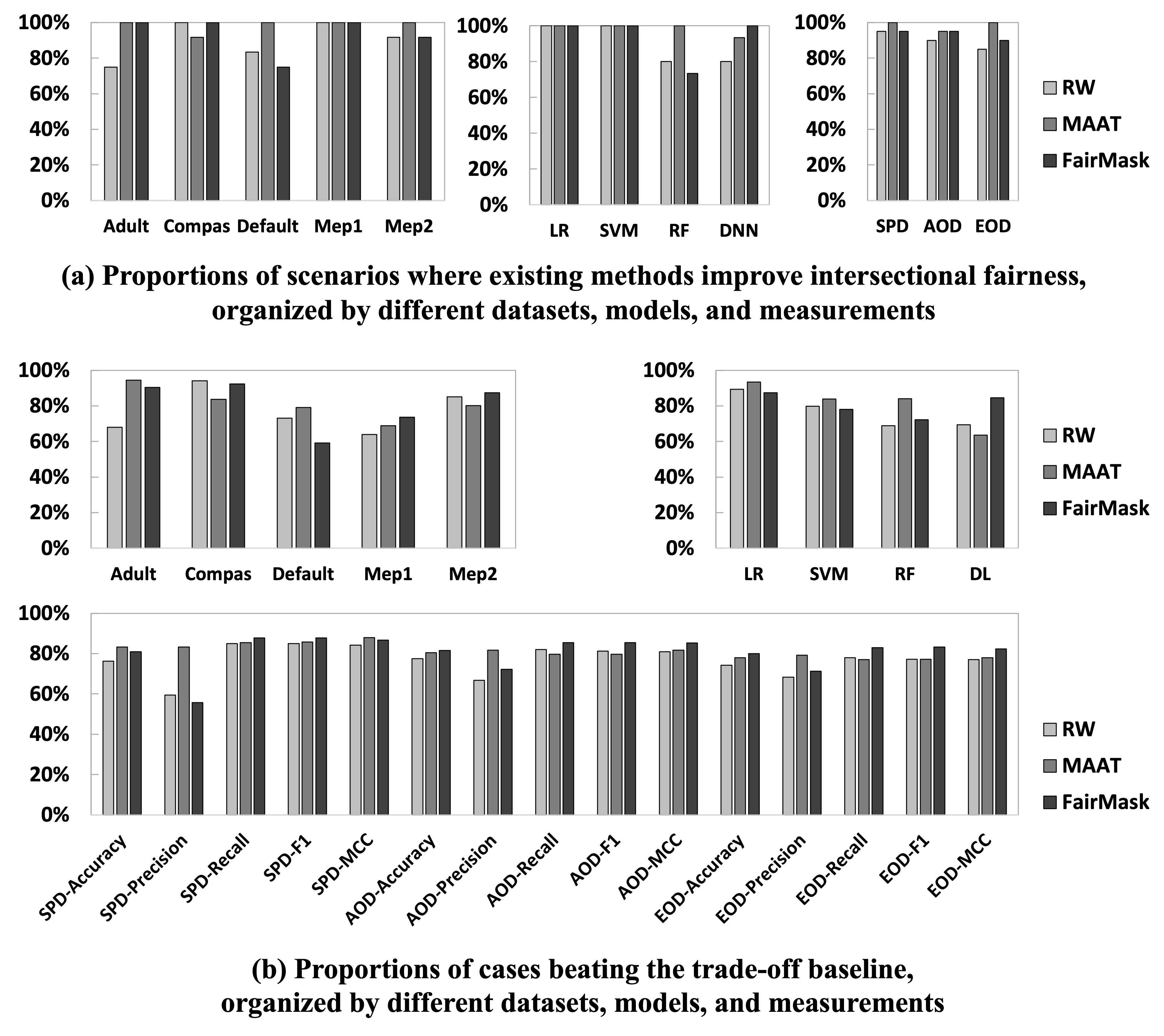}
  \caption{(RQ4) Effectiveness in intersectional fairness improvement and fairness-performance trade-off of the best three methods identified in this study (i.e., RW, MAAT, and FairMask) across various datasets, models, and measurements. We observe that it is challenging for these methods to achieve a good fairness-performance trade-off for imbalanced datasets and precision-critical applications.}
  \label{figff} 
\end{figure*}

Due to the page limit, we show only the results of the top three methods identified in RQ2 and RQ3 (i.e., RW, MAAT, and FairMask) in Figure~\ref{figff}, and the results for all methods can be found in our repository~\cite{githublink}.

As shown in Figure \ref{figff}(a), for each dataset and each model, at least one of the methods RW, MAAT, and FairMask can improve intersectional fairness in 100\% of scenarios. However, regarding the fairness measurements, all three methods cannot do so for AOD. It is reasonable since the AOD fairness is more complex and difficult to satisfy than SPD and EOD, as demonstrated in previous work~\cite{Dabs220703277}.

Figure \ref{figff}(b) reveals that these methods tend to achieve worse fairness-performance trade-offs on imbalanced datasets compared to balanced datasets. Specifically, from Table \ref{dataset_info}, we find that the majority class in the Adult, Compas, Default, Mep1, and Mep2 datasets accounts for 76.1\%, 54.9\%, 77.9\%, 82.8\%, and 83.2\%, respectively. Among these datasets, Compas, being the most balanced, exhibits the best fairness-performance trade-off results. This observation is expected since the classification on balanced datasets is generally considered easier than on imbalanced ones \cite{ramyachitra2014imbalanced}, making it relatively easier for existing methods to retain performance while improving fairness on such datasets. In addition, our findings indicate that achieving a good trade-off between fairness and precision is overall more challenging for existing methods compared to the trade-off between fairness and other performance metrics. 

\finding{It is challenging for fairness improvement methods to achieve good fairness-performance trade-offs for imbalanced datasets and applications where precision matters when dealing with multiple protected attributes.}

\section{Implications}
\noindent \textbf{Implications for software engineers:} \textbf{1)}~There is a substantial risk of inadvertently exacerbating unfairness for unconsidered protected attributes and violating anti-discrimination laws when software engineers focus on certain protected attributes. This is due to the presence of a noteworthy trade-off between fairness across different protected attributes observed in our study.
If the trade-off comes simply because the data is skewed thus creating `artificial contention' between protected attributes, it can be corrected by software engineers, as a type of fairness bug. Otherwise, if it is inherent to the problem that there is a trade-off between the fairness regarding different protected attributes, the competing fairness requirements raise issues of negotiation, mediation, and conflict resolution for engineers.
\textbf{2)} We have compared 11 state-of-the-art fairness improvement methods when dealing with multiple protected attributes based on several different metrics. The results offer valuable insights and references for software engineers when they select fairness improvement methods that address multiple protected attributes in line with their specific objectives, thereby mitigating legal risks associated with software discrimination. For example, the results of RQ2 reveal that when faced with limited access to training data, the EOP method emerges as a viable choice for improving intersectional fairness. Conversely, MAAT can be a suitable option while having access to training data. 

\noindent \textbf{Implications for policy makers:} Despite many laws and regulations seeking to protect multiple attributes simultaneously~\cite{GhoshGR21,usequal}, our findings reveal that fairness objectives for protected attributes such as sex and race may compete with each other. As a result, expecting software systems to perfectly satisfy these competing fairness objectives under a single law or regulation can be unrealistic. To achieve a balanced approach towards fairness in software systems, policy makers and legislative bodies should carefully consider these competing fairness considerations when formulating laws and regulations.

\noindent \textbf{Implications for researchers:} 
\textbf{1)} There is a potential risk associated with the common research practice of focusing on one protected attribute at a time, as fairness improvement methods can significantly impact fairness regarding unconsidered protected attributes (RQ1). This emphasizes the importance of considering multiple protected attributes, not only in real-world applications, but also as a crucial objective in research. Researchers should be mindful of the potential consequences of neglecting the impact on unconsidered protected attributes and strive to broaden the scope of their investigations to encompass multiple dimensions of fairness. 
\textbf{2)}~Considering the well-known fairness-performance trade-off and the trade-off between fairness regarding different protected attributes observed in our study (RQ1), researchers have the opportunity to develop multi-objective optimization techniques that address both these trade-offs simultaneously. 
\textbf{3)} Researchers can prioritize proposing post-processing fairness improvement techniques for tackling multiple protected attributes. This focus is driven by the finding that RW, MAAT, and FairMask are the most effective methods for enhancing intersectional fairness (RQ2), but they all require access to training data, posing challenges in real-world fairness-related applications due to concerns about releasing sensitive personal information. In contrast, EOP, the top-performing post-processing method that does not require such access, achieves intersectional fairness improvement in 18.3\% fewer scenarios (RQ2). 
\textbf{4)} Researchers should include F1-score and MCC in their evaluations when dealing with multiple protected attributes, moving beyond sole reliance on accuracy, as commonly observed in existing fairness research \cite{icseZhangH21,sigsoftHortZSH21,icdmCaldersKP09,datamineCaldersV10,mfcpaper,kddFeldmanFMSV15,rewpaper,icdmKamiranCP10,isciKamiranMKZ18,aistatsZafarVGG17}. It is because fairness improvements can have a significant impact on F1-score and MCC when considering multiple protected attributes (RQ3). F1-score and MCC's wide adoption in real-world applications further emphasizes the importance \cite{Dabs220703277,isstaMoussaS22}.
\textbf{5)}~Researchers can design novel methods specifically tailored to optimize the fairness-performance trade-off for imbalanced datasets and precision-critical applications, because existing methods may not suffice under such circumstances~(RQ4). This is important especially considering that these circumstances are common in the real-world applications~\cite{kotsiantis2006handling,kanei2019precise}.

\section{Threats to Validity}
\noindent \textbf{Datasets:} Due to the lack of public availability of datasets across all domains with fairness issues, we use five widely-adopted datasets that cover common domains frequently explored in the fairness literature. However, it is important to note that these widely-adopted datasets can have potential limitations \cite{ding2021retiring}, which may affect the validity of our findings. In addition, regarding protected attributes, we consider only sex, race, and age, which are the most widely-studied ones in the fairness literature \cite{soremekun2022software}. In the future, one could replicate this study with more datasets and more protected attributes.

\noindent \textbf{ML models:} To mitigate potential concerns regarding the selection of ML models, we have carefully chosen representative models for our study. Our selection includes both traditional ML models such as LR, RF, and SVM, as well as DNN. LR, RF, and SVM have been widely adopted in decision-making scenarios of social significance where fairness is a critical factor, as supported by existing research \cite{sigsoftChenZSH22} and a recent official report from the UK government \cite{ukreport}. Moreover, DNN is increasingly adopted in the fairness literature due to their expanding applications in decision-making contexts~\cite{sigsoftZhang022,Dabs220703277,sigsoftTaoSHF022,icseZhengCD0CJW0C22}. 

\noindent \textbf{Fairness improvement methods:} In recent years, the significance of fairness has gained considerable attention, resulting in an increasing number of fairness improvement methods. Given the extensive range of methods available, it is challenging to incorporate all of them in our study. To address this limitation, we choose 11 representative methods that have been recognized as state-of-the-art in the literature \cite{sigsoftChenZSH22, sigsoftZhang022, fairsmotepaper, fairmaskpaper}. 
While we have considered a wide range of fairness improvement methods that can be applied to different phases of the machine learning pipeline, we acknowledge that, in practice, they are not always applicable, given the constraints of the data sources and the application domain.

\noindent \textbf{Evaluation metrics:} 
Fairness metrics have been increasingly emerging in the literature.
It is impractical to incorporate all of these metrics in our study. To address this limitation, we have followed previous studies \cite{sigsoftChenZSH22} to use three fairness metrics that have gained significant adoption in the literature. Similarly, for performance evaluation, we have used the most widely-adopted metrics for ML classification \cite{sigsoftChenZSH22}.
We have employed a comprehensive set of 15 fairness-performance measurements, which is the most extensive range used in the literature.

\section{Related Work}
Researchers have made significant efforts to address unfairness issues in ML software by proposing various fairness improvement methods. For instance, IBM has launched the AIF360 toolkit that integrates cutting-edge fairness improvement methods \cite{corrbs181001943}, such as Reweighting \cite{rewpaper}, Prejudice Remover \cite{PRpaper}, and Equalized Odds Processing \cite{EOpaper}. These methods can be categorized into pre-, in-, and post-processing methods, which respectively optimize training data, the learning process, and decision outputs to improve fairness~\cite{Dabs220703277}.
While a plethora of fairness improvement methods have been proposed, the majority of them primarily concentrate on addressing individual protected attributes, as emphasized in recent work \cite{fairsmotepaper,fairmaskpaper,sigsoftChenZSH22,GhoshGR21}.

With the increasing number of fairness improvement methods, previous studies have aimed to empirically evaluate and compare existing methods. For instance, Biswas and Rajan \cite{biswas2020machine} assessed seven fairness improvement methods using ML models gathered from a crowd-sourced platform, analyzing the resulting fairness outcomes and their impact on performance. Hort et al.~\cite{sigsoftHortZSH21} introduced Fairea, a benchmarking tool that provides a unified baseline for evaluating the fairness-performance trade-off obtained by different methods. Chen et al. \cite{Dabs220703277} used Fairea to conduct a comprehensive empirical study of state-of-the-art fairness improvement methods. However, all these evaluations are limited to tasks involving a single protected attribute at a time.

Recent SE studies have presented methods capable of handling multiple protected attributes simultaneously \cite{sigsoftChenZSH22,fairsmotepaper,fairmaskpaper}. 
However, the systematic comparison of these methods remains understudied. 
Specifically, when evaluating their method for dealing with multiple protected attributes, Chakraborty et al.~\cite{fairsmotepaper} did not employ any method for comparison; Chen et al.~\cite{sigsoftChenZSH22} and Peng et al.~\cite{fairmaskpaper} compared the proposed methods with only the one proposed by Chakraborty et al.~\cite{fairsmotepaper}. Additionally, the effectiveness of these methods in improving intersectional fairness was not evaluated in previous work. Recently, Zhang and Sun \cite{sigsoftZhang022} adapted fairness improvement methods previously proposed in the ML community so that they can handle multiple protected attributes. However, they did not compare these methods with the recent ones proposed by the SE community~\cite{fairsmotepaper,sigsoftChenZSH22,fairmaskpaper}, and they used SPD as the only group fairness metric and accuracy as the only performance metric for evaluation. In this paper, we systematically study the effectiveness of 11 state-of-the-art fairness improvement methods (covering methods from both ML and SE communities) in improving intersectional fairness with multiple widely-adopted fairness metrics. We also investigate the fairness-performance trade-off achieved by these methods in the context of multiple protected attributes using 15 fairness-performance measurements.

\section{Conclusion}
This paper presents an extensive study of fairness improvement with multiple protected attributes. We systematically study 11 state-of-the-art fairness improvement methods from the literature, on widely-adopted benchmark datasets, ML models, performance metrics, and fairness metrics. We uncover the potential trade-off between fairness regarding different protected attributes and find that the correlation between the attributes can be a possible reason. We also explore the influence on performance when improving fairness for multiple protected attributes. Moreover, we benchmark existing methods and compare their effectiveness in improving intersectional fairness and achieving the trade-off between intersectional fairness and performance. The results provide actionable implications for researchers, software engineers, and policy makers.


\section{Data Availability}
We have made the code and data used in this paper publicly accessible~\cite{githublink}.

\begin{acks}
Zhenpeng Chen, Federica Sarro, and Mark Harman are supported by the ERC Advanced Grant No.741278 (EPIC: Evolutionary Program Improvement Collaborators). Jie M. Zhang is supported by the UKRI Trustworthy Autonomous Systems Node in Verifiability, with Grant Award Reference EP/V026801/2.
\end{acks}

\balance
\bibliographystyle{ACM-Reference-Format}
\bibliography{fairness-bib}


\begin{thebibliography}{67}


\ifx \showCODEN    \undefined \def \showCODEN     #1{\unskip}     \fi
\ifx \showDOI      \undefined \def \showDOI       #1{#1}\fi
\ifx \showISBNx    \undefined \def \showISBNx     #1{\unskip}     \fi
\ifx \showISBNxiii \undefined \def \showISBNxiii  #1{\unskip}     \fi
\ifx \showISSN     \undefined \def \showISSN      #1{\unskip}     \fi
\ifx \showLCCN     \undefined \def \showLCCN      #1{\unskip}     \fi
\ifx \shownote     \undefined \def \shownote      #1{#1}          \fi
\ifx \showarticletitle \undefined \def \showarticletitle #1{#1}   \fi
\ifx \showURL      \undefined \def \showURL       {\relax}        \fi
\providecommand\bibfield[2]{#2}
\providecommand\bibinfo[2]{#2}
\providecommand\natexlab[1]{#1}
\providecommand\showeprint[2][]{arXiv:#2}

\bibitem[adu(1996)]%
        {adultdata}
 \bibinfo{year}{1996}\natexlab{}.
\newblock \bibinfo{title}{The Adult Census Income dataset}.
\newblock \bibinfo{howpublished}{\url{https://archive.ics.uci.edu/ml/datasets/adult}}.
\newblock


\bibitem[use(2003)]%
        {usequal}
 \bibinfo{year}{2003}\natexlab{}.
\newblock \bibinfo{title}{U.S. Equal Employment Opportunity Commission}.
\newblock \bibinfo{howpublished}{\url{https://www.eeoc.gov/initiatives/e-race/significant-eeoc-racecolor-casescovering-private-and-federal-sectors\#intersectional}}.
\newblock


\bibitem[mep(2015)]%
        {mep15data}
 \bibinfo{year}{2015}\natexlab{}.
\newblock \bibinfo{title}{The Mep15 dataset}.
\newblock \bibinfo{howpublished}{\url{https://meps.ahrq.gov/mepsweb/data_stats/download_data_files_detail.jsp?cboPufNumber=HC-181}}.
\newblock


\bibitem[com(2016)]%
        {compasdata}
 \bibinfo{year}{2016}\natexlab{}.
\newblock \bibinfo{title}{The Compas dataset}.
\newblock \bibinfo{howpublished}{\url{https://github.com/propublica/compas-analysis}}.
\newblock


\bibitem[def(2016)]%
        {defaultdata}
 \bibinfo{year}{2016}\natexlab{}.
\newblock \bibinfo{title}{The Default dataset}.
\newblock \bibinfo{howpublished}{\url{https://archive.ics.uci.edu/ml/datasets/default+of+credit+card+clients}}.
\newblock


\bibitem[fai(2016)]%
        {faircase}
 \bibinfo{year}{2016}\natexlab{}.
\newblock \bibinfo{title}{Machine bias}.
\newblock \bibinfo{howpublished}{\url{https://www.propublica.org/article/machine-bias-risk-assessments-in-criminal-sentencing}}.
\newblock


\bibitem[mep(2016)]%
        {mep16data}
 \bibinfo{year}{2016}\natexlab{}.
\newblock \bibinfo{title}{The Mep16 dataset}.
\newblock \bibinfo{howpublished}{\url{https://meps.ahrq.gov/mepsweb/data_stats/download_data_files_detail.jsp?cboPufNumber=HC-192}}.
\newblock


\bibitem[fai(2018)]%
        {faircase2}
 \bibinfo{year}{2018}\natexlab{}.
\newblock \bibinfo{title}{Study finds gender and skin-type bias in commercial artificial-intelligence systems}.
\newblock \bibinfo{howpublished}{\url{https://news.mit.edu/2018/study-finds-gender-skin-type-bias-artificial-intelligence-systems-0212}}.
\newblock


\bibitem[ukr(2020)]%
        {ukreport}
 \bibinfo{year}{2020}\natexlab{}.
\newblock \bibinfo{title}{Review into bias in algorithmic decision-making}.
\newblock \bibinfo{howpublished}{\url{https://www.gov.uk/government/publications/cdei-publishes-review-into-bias-in-algorithmic-decision-making/main-report-cdei-review-into-bias-in-algorithmic-decision-making}}.
\newblock


\bibitem[fai(2021)]%
        {faircase1}
 \bibinfo{year}{2021}\natexlab{}.
\newblock \bibinfo{title}{When good algorithms go sexist: Why and how to advance AI gender equity}.
\newblock \bibinfo{howpublished}{\url{https://ssir.org/articles/entry/when_good_algorithms_go_sexist_why_and_how_to_advance_ai_gender_equity}}.
\newblock


\bibitem[git(2024)]%
        {githublink}
 \bibinfo{year}{2024}\natexlab{}.
\newblock \bibinfo{title}{Replication package}.
\newblock \bibinfo{howpublished}{\url{https://doi.org/10.6084/m9.figshare.24943590.v3}}.
\newblock


\bibitem[Andreeva et~al\mbox{.}(2004)]%
        {andreeva2004impact}
\bibfield{author}{\bibinfo{person}{Galina Andreeva}, \bibinfo{person}{Jake Ansell}, {and} \bibinfo{person}{Jonathan Crook}.} \bibinfo{year}{2004}\natexlab{}.
\newblock \showarticletitle{Impact of anti-discrimination laws on credit scoring}.
\newblock \bibinfo{journal}{\emph{Journal of Financial Services Marketing}}  \bibinfo{volume}{9} (\bibinfo{year}{2004}), \bibinfo{pages}{22--33}.
\newblock


\bibitem[Bellamy et~al\mbox{.}(2019)]%
        {corrbs181001943}
\bibfield{author}{\bibinfo{person}{Rachel K.~E. Bellamy}, \bibinfo{person}{Kuntal Dey}, \bibinfo{person}{Michael Hind}, \bibinfo{person}{Samuel~C. Hoffman}, \bibinfo{person}{Stephanie Houde}, \bibinfo{person}{Kalapriya Kannan}, \bibinfo{person}{Pranay Lohia}, \bibinfo{person}{Jacquelyn Martino}, \bibinfo{person}{Sameep Mehta}, \bibinfo{person}{Aleksandra Mojsilovic}, \bibinfo{person}{Seema Nagar}, \bibinfo{person}{Karthikeyan~Natesan Ramamurthy}, \bibinfo{person}{John~T. Richards}, \bibinfo{person}{Diptikalyan Saha}, \bibinfo{person}{Prasanna Sattigeri}, \bibinfo{person}{Moninder Singh}, \bibinfo{person}{Kush~R. Varshney}, {and} \bibinfo{person}{Yunfeng Zhang}.} \bibinfo{year}{2019}\natexlab{}.
\newblock \showarticletitle{{AI} Fairness 360: An extensible toolkit for detecting and mitigating algorithmic bias}.
\newblock \bibinfo{journal}{\emph{IBM Journal of Research and Development}} \bibinfo{volume}{63}, \bibinfo{number}{4/5} (\bibinfo{year}{2019}), \bibinfo{pages}{4:1--4:15}.
\newblock


\bibitem[Bennin et~al\mbox{.}(2017)]%
        {esemBenninKMPM17}
\bibfield{author}{\bibinfo{person}{Kwabena~Ebo Bennin}, \bibinfo{person}{Jacky Keung}, \bibinfo{person}{Akito Monden}, \bibinfo{person}{Passakorn Phannachitta}, {and} \bibinfo{person}{Solomon Mensah}.} \bibinfo{year}{2017}\natexlab{}.
\newblock \showarticletitle{The significant effects of data sampling approaches on software defect prioritization and classification}. In \bibinfo{booktitle}{\emph{Proceedings of the 2017 {ACM/IEEE} International Symposium on Empirical Software Engineering and Measurement, {ESEM} 2017}}. \bibinfo{pages}{364--373}.
\newblock


\bibitem[Berk et~al\mbox{.}(2021)]%
        {berk2021fairness}
\bibfield{author}{\bibinfo{person}{Richard Berk}, \bibinfo{person}{Hoda Heidari}, \bibinfo{person}{Shahin Jabbari}, \bibinfo{person}{Michael Kearns}, {and} \bibinfo{person}{Aaron Roth}.} \bibinfo{year}{2021}\natexlab{}.
\newblock \showarticletitle{Fairness in criminal justice risk assessments: The state of the art}.
\newblock \bibinfo{journal}{\emph{Sociological Methods \& Research}} \bibinfo{volume}{50}, \bibinfo{number}{1} (\bibinfo{year}{2021}), \bibinfo{pages}{3--44}.
\newblock


\bibitem[Biswas and Rajan(2020)]%
        {biswas2020machine}
\bibfield{author}{\bibinfo{person}{Sumon Biswas} {and} \bibinfo{person}{Hridesh Rajan}.} \bibinfo{year}{2020}\natexlab{}.
\newblock \showarticletitle{Do the machine learning models on a crowd sourced platform exhibit bias? An empirical study on model fairness}. In \bibinfo{booktitle}{\emph{Proceedings of the 28th ACM Joint Meeting on European Software Engineering Conference and Symposium on the Foundations of Software Engineering, ESEC/FSE 2020}}. \bibinfo{pages}{642--653}.
\newblock


\bibitem[Biswas and Rajan(2021)]%
        {sigsoftBiswasR21}
\bibfield{author}{\bibinfo{person}{Sumon Biswas} {and} \bibinfo{person}{Hridesh Rajan}.} \bibinfo{year}{2021}\natexlab{}.
\newblock \showarticletitle{Fair preprocessing: Towards understanding compositional fairness of data transformers in machine learning pipeline}. In \bibinfo{booktitle}{\emph{Proceedings of the 29th {ACM} Joint European Software Engineering Conference and Symposium on the Foundations of Software Engineering, {ESEC/FSE} 2021}}. \bibinfo{pages}{981--993}.
\newblock


\bibitem[Biswas and Rajan(2023)]%
        {icseBiswasR23}
\bibfield{author}{\bibinfo{person}{Sumon Biswas} {and} \bibinfo{person}{Hridesh Rajan}.} \bibinfo{year}{2023}\natexlab{}.
\newblock \showarticletitle{Fairify: Fairness verification of neural networks}. In \bibinfo{booktitle}{\emph{Proceedings of the 45th {IEEE/ACM} International Conference on Software Engineering, {ICSE} 2023}}. \bibinfo{pages}{1546--1558}.
\newblock


\bibitem[Calders et~al\mbox{.}(2009)]%
        {icdmCaldersKP09}
\bibfield{author}{\bibinfo{person}{Toon Calders}, \bibinfo{person}{Faisal Kamiran}, {and} \bibinfo{person}{Mykola Pechenizkiy}.} \bibinfo{year}{2009}\natexlab{}.
\newblock \showarticletitle{Building classifiers with independency constraints}. In \bibinfo{booktitle}{\emph{Proceedings of the 2009 {IEEE} International Conference on Data Mining}}. \bibinfo{pages}{13--18}.
\newblock


\bibitem[Calders and Verwer(2010)]%
        {datamineCaldersV10}
\bibfield{author}{\bibinfo{person}{Toon Calders} {and} \bibinfo{person}{Sicco Verwer}.} \bibinfo{year}{2010}\natexlab{}.
\newblock \showarticletitle{Three naive Bayes approaches for discrimination-free classification}.
\newblock \bibinfo{journal}{\emph{Data Mining and Knowledge Discovery}} \bibinfo{volume}{21}, \bibinfo{number}{2} (\bibinfo{year}{2010}), \bibinfo{pages}{277--292}.
\newblock


\bibitem[Celis et~al\mbox{.}(2019)]%
        {mfcpaper}
\bibfield{author}{\bibinfo{person}{L.~Elisa Celis}, \bibinfo{person}{Lingxiao Huang}, \bibinfo{person}{Vijay Keswani}, {and} \bibinfo{person}{Nisheeth~K. Vishnoi}.} \bibinfo{year}{2019}\natexlab{}.
\newblock \showarticletitle{Classification with fairness constraints: A meta-algorithm wit provable guarantees}. In \bibinfo{booktitle}{\emph{Proceedings of the Conference on Fairness, Accountability, and Transparency, FAT* 2019}}. \bibinfo{pages}{319--328}.
\newblock


\bibitem[Chakraborty et~al\mbox{.}(2021)]%
        {fairsmotepaper}
\bibfield{author}{\bibinfo{person}{Joymallya Chakraborty}, \bibinfo{person}{Suvodeep Majumder}, {and} \bibinfo{person}{Tim Menzies}.} \bibinfo{year}{2021}\natexlab{}.
\newblock \showarticletitle{Bias in machine learning software: Why? How? What to do?}. In \bibinfo{booktitle}{\emph{Proceedings of the 29th {ACM} Joint European Software Engineering Conference and Symposium on the Foundations of Software Engineering, {ESEC/FSE} 2021}}. \bibinfo{pages}{429--440}.
\newblock


\bibitem[Chakraborty et~al\mbox{.}(2020)]%
        {fairwaypaper}
\bibfield{author}{\bibinfo{person}{Joymallya Chakraborty}, \bibinfo{person}{Suvodeep Majumder}, \bibinfo{person}{Zhe Yu}, {and} \bibinfo{person}{Tim Menzies}.} \bibinfo{year}{2020}\natexlab{}.
\newblock \showarticletitle{Fairway: A way to build fair {ML} software}. In \bibinfo{booktitle}{\emph{Proceedings of the 28th {ACM} Joint European Software Engineering Conference and Symposium on the Foundations of Software Engineering, {ESEC/FSE} 2020}}. \bibinfo{pages}{654--665}.
\newblock


\bibitem[Chen et~al\mbox{.}(2022a)]%
        {Dabs220710223}
\bibfield{author}{\bibinfo{person}{Zhenpeng Chen}, \bibinfo{person}{Jie~M. Zhang}, \bibinfo{person}{Max Hort}, \bibinfo{person}{Federica Sarro}, {and} \bibinfo{person}{Mark Harman}.} \bibinfo{year}{2022}\natexlab{a}.
\newblock \showarticletitle{Fairness testing: A comprehensive survey and analysis of trends}.
\newblock \bibinfo{journal}{\emph{CoRR}}  \bibinfo{volume}{abs/2207.10223} (\bibinfo{year}{2022}).
\newblock


\bibitem[Chen et~al\mbox{.}(2022b)]%
        {sigsoftChenZSH22}
\bibfield{author}{\bibinfo{person}{Zhenpeng Chen}, \bibinfo{person}{Jie~M. Zhang}, \bibinfo{person}{Federica Sarro}, {and} \bibinfo{person}{Mark Harman}.} \bibinfo{year}{2022}\natexlab{b}.
\newblock \showarticletitle{{MAAT:} a novel ensemble approach to addressing fairness and performance bugs for machine learning software}. In \bibinfo{booktitle}{\emph{Proceedings of the 30th {ACM} Joint European Software Engineering Conference and Symposium on the Foundations of Software Engineering, {ESEC/FSE} 2022}}. \bibinfo{pages}{1122--1134}.
\newblock


\bibitem[Chen et~al\mbox{.}(2023)]%
        {Dabs220703277}
\bibfield{author}{\bibinfo{person}{Zhenpeng Chen}, \bibinfo{person}{Jie~M. Zhang}, \bibinfo{person}{Federica Sarro}, {and} \bibinfo{person}{Mark Harman}.} \bibinfo{year}{2023}\natexlab{}.
\newblock \showarticletitle{A comprehensive empirical study of bias mitigation methods for machine learning classifiers}.
\newblock \bibinfo{journal}{\emph{{ACM} Transactions on Software Engineering and Methodology}} \bibinfo{volume}{32}, \bibinfo{number}{4} (\bibinfo{year}{2023}), \bibinfo{pages}{106:1--106:30}.
\newblock


\bibitem[Corbett{-}Davies et~al\mbox{.}(2017)]%
        {CorbettDaviesP17}
\bibfield{author}{\bibinfo{person}{Sam Corbett{-}Davies}, \bibinfo{person}{Emma Pierson}, \bibinfo{person}{Avi Feller}, \bibinfo{person}{Sharad Goel}, {and} \bibinfo{person}{Aziz Huq}.} \bibinfo{year}{2017}\natexlab{}.
\newblock \showarticletitle{Algorithmic decision making and the cost of fairness}. In \bibinfo{booktitle}{\emph{Proceedings of the 23rd {ACM} {SIGKDD} International Conference on Knowledge Discovery and Data Mining, KDD 2017}}. \bibinfo{pages}{797--806}.
\newblock


\bibitem[Crenshaw(1989)]%
        {crenshaw1989demarginalizing}
\bibfield{author}{\bibinfo{person}{Kimberle Crenshaw}.} \bibinfo{year}{1989}\natexlab{}.
\newblock \showarticletitle{Demarginalizing the intersection of race and sex: A black feminist critique of antidiscrimination doctrine, deminist theory and antiracist politics}.
\newblock \bibinfo{journal}{\emph{Feminist Legal Theories}} (\bibinfo{year}{1989}), \bibinfo{pages}{139--167}.
\newblock


\bibitem[Ding et~al\mbox{.}(2021)]%
        {ding2021retiring}
\bibfield{author}{\bibinfo{person}{Frances Ding}, \bibinfo{person}{Moritz Hardt}, \bibinfo{person}{John Miller}, {and} \bibinfo{person}{Ludwig Schmidt}.} \bibinfo{year}{2021}\natexlab{}.
\newblock \showarticletitle{Retiring adult: New datasets for fair machine learning}.
\newblock \bibinfo{journal}{\emph{Advances in neural information processing systems}}  \bibinfo{volume}{34} (\bibinfo{year}{2021}), \bibinfo{pages}{6478--6490}.
\newblock


\bibitem[Feldman et~al\mbox{.}(2015)]%
        {kddFeldmanFMSV15}
\bibfield{author}{\bibinfo{person}{Michael Feldman}, \bibinfo{person}{Sorelle~A. Friedler}, \bibinfo{person}{John Moeller}, \bibinfo{person}{Carlos Scheidegger}, {and} \bibinfo{person}{Suresh Venkatasubramanian}.} \bibinfo{year}{2015}\natexlab{}.
\newblock \showarticletitle{Certifying and removing disparate impact}. In \bibinfo{booktitle}{\emph{Proceedings of the 21th {ACM} {SIGKDD} International Conference on Knowledge Discovery and Data Mining}}. \bibinfo{pages}{259--268}.
\newblock


\bibitem[Finkelstein et~al\mbox{.}(2008)]%
        {FinkelsteinHMRZ08}
\bibfield{author}{\bibinfo{person}{Anthony Finkelstein}, \bibinfo{person}{Mark Harman}, \bibinfo{person}{S.~Afshin Mansouri}, \bibinfo{person}{Jian Ren}, {and} \bibinfo{person}{Yuanyuan Zhang}.} \bibinfo{year}{2008}\natexlab{}.
\newblock \showarticletitle{``Fairness analysis'' in requirements assignments}. In \bibinfo{booktitle}{\emph{Proceedings of the 16th {IEEE} International Requirements Engineering Conference, {RE} 2008}}. \bibinfo{pages}{115--124}.
\newblock


\bibitem[Foulds et~al\mbox{.}(2020)]%
        {icdeFouldsIKP20}
\bibfield{author}{\bibinfo{person}{James~R. Foulds}, \bibinfo{person}{Rashidul Islam}, \bibinfo{person}{Kamrun~Naher Keya}, {and} \bibinfo{person}{Shimei Pan}.} \bibinfo{year}{2020}\natexlab{}.
\newblock \showarticletitle{An intersectional definition of fairness}. In \bibinfo{booktitle}{\emph{Proceedings of the 36th {IEEE} International Conference on Data Engineering, {ICDE} 2020}}. \bibinfo{pages}{1918--1921}.
\newblock


\bibitem[Ghosh et~al\mbox{.}(2021)]%
        {GhoshGR21}
\bibfield{author}{\bibinfo{person}{Avijit Ghosh}, \bibinfo{person}{Lea Genuit}, {and} \bibinfo{person}{Mary Reagan}.} \bibinfo{year}{2021}\natexlab{}.
\newblock \showarticletitle{Characterizing intersectional group fairness with worst-case comparisons}. In \bibinfo{booktitle}{\emph{Proceedings of the Artificial Intelligence Diversity, Belonging, Equity, and Inclusion, {AIDBEI} 2021}}. \bibinfo{pages}{22--34}.
\newblock


\bibitem[Gohar et~al\mbox{.}(2023)]%
        {icseGoharBR23}
\bibfield{author}{\bibinfo{person}{Usman Gohar}, \bibinfo{person}{Sumon Biswas}, {and} \bibinfo{person}{Hridesh Rajan}.} \bibinfo{year}{2023}\natexlab{}.
\newblock \showarticletitle{Towards understanding fairness and its composition in ensemble machine learning}. In \bibinfo{booktitle}{\emph{Proceedings of the 45th {IEEE/ACM} International Conference on Software Engineering, {ICSE} 2023}}. \bibinfo{pages}{1533--1545}.
\newblock


\bibitem[Hardt et~al\mbox{.}(2016)]%
        {EOpaper}
\bibfield{author}{\bibinfo{person}{Moritz Hardt}, \bibinfo{person}{Eric Price}, {and} \bibinfo{person}{Nati Srebro}.} \bibinfo{year}{2016}\natexlab{}.
\newblock \showarticletitle{Equality of opportunity in supervised learning}. In \bibinfo{booktitle}{\emph{Proceedings of the Annual Conference on Neural Information Processing Systems 2016, NIPS 2016}}. \bibinfo{pages}{3315--3323}.
\newblock


\bibitem[Hort et~al\mbox{.}(2023)]%
        {DBcorrabs220707068}
\bibfield{author}{\bibinfo{person}{Max Hort}, \bibinfo{person}{Zhenpeng Chen}, \bibinfo{person}{Jie~M. Zhang}, \bibinfo{person}{Mark Harman}, {and} \bibinfo{person}{Federica Sarro}.} \bibinfo{year}{2023}\natexlab{}.
\newblock \showarticletitle{Bias mitigation for machine learning classifiers: {A} comprehensive survey}.
\newblock \bibinfo{journal}{\emph{ACM Journal on Responsible Computing}} (\bibinfo{year}{2023}).
\newblock


\bibitem[Hort and Sarro(2021)]%
        {maxasefairness}
\bibfield{author}{\bibinfo{person}{Max Hort} {and} \bibinfo{person}{Federica Sarro}.} \bibinfo{year}{2021}\natexlab{}.
\newblock \showarticletitle{Did you do your homework? Raising awareness on software fairness and discrimination}. In \bibinfo{booktitle}{\emph{Proceedings of the 36th IEEE/ACM International Conference on Automated Software Engineering, ASE 2021}}. \bibinfo{pages}{1322--1326}.
\newblock


\bibitem[Hort et~al\mbox{.}(2021)]%
        {sigsoftHortZSH21}
\bibfield{author}{\bibinfo{person}{Max Hort}, \bibinfo{person}{Jie~M. Zhang}, \bibinfo{person}{Federica Sarro}, {and} \bibinfo{person}{Mark Harman}.} \bibinfo{year}{2021}\natexlab{}.
\newblock \showarticletitle{Fairea: A model behaviour mutation approach to benchmarking bias mitigation methods}. In \bibinfo{booktitle}{\emph{Proceedings of the 29th {ACM} Joint European Software Engineering Conference and Symposium on the Foundations of Software Engineering, ESEC/FSE 2021}}. \bibinfo{pages}{994--1006}.
\newblock


\bibitem[Kamiran and Calders(2011)]%
        {rewpaper}
\bibfield{author}{\bibinfo{person}{Faisal Kamiran} {and} \bibinfo{person}{Toon Calders}.} \bibinfo{year}{2011}\natexlab{}.
\newblock \showarticletitle{Data preprocessing techniques for classification without discrimination}.
\newblock \bibinfo{journal}{\emph{Knowledge and Information Systems}} \bibinfo{volume}{33}, \bibinfo{number}{1} (\bibinfo{year}{2011}), \bibinfo{pages}{1--33}.
\newblock


\bibitem[Kamiran et~al\mbox{.}(2010)]%
        {icdmKamiranCP10}
\bibfield{author}{\bibinfo{person}{Faisal Kamiran}, \bibinfo{person}{Toon Calders}, {and} \bibinfo{person}{Mykola Pechenizkiy}.} \bibinfo{year}{2010}\natexlab{}.
\newblock \showarticletitle{Discrimination aware decision tree learning}. In \bibinfo{booktitle}{\emph{Proceedings of the 10th {IEEE} International Conference on Data Mining, ICDM 2010}}. \bibinfo{pages}{869--874}.
\newblock


\bibitem[Kamiran et~al\mbox{.}(2012)]%
        {ROCpaper}
\bibfield{author}{\bibinfo{person}{Faisal Kamiran}, \bibinfo{person}{Asim Karim}, {and} \bibinfo{person}{Xiangliang Zhang}.} \bibinfo{year}{2012}\natexlab{}.
\newblock \showarticletitle{Decision theory for discrimination-aware classification}. In \bibinfo{booktitle}{\emph{Proceedings of the 12th {IEEE} International Conference on Data Mining, {ICDM} 2012}}. \bibinfo{pages}{924--929}.
\newblock


\bibitem[Kamiran et~al\mbox{.}(2018)]%
        {isciKamiranMKZ18}
\bibfield{author}{\bibinfo{person}{Faisal Kamiran}, \bibinfo{person}{Sameen Mansha}, \bibinfo{person}{Asim Karim}, {and} \bibinfo{person}{Xiangliang Zhang}.} \bibinfo{year}{2018}\natexlab{}.
\newblock \showarticletitle{Exploiting reject option in classification for social discrimination control}.
\newblock \bibinfo{journal}{\emph{Information Science}}  \bibinfo{volume}{425} (\bibinfo{year}{2018}), \bibinfo{pages}{18--33}.
\newblock


\bibitem[Kamishima et~al\mbox{.}(2012)]%
        {PRpaper}
\bibfield{author}{\bibinfo{person}{Toshihiro Kamishima}, \bibinfo{person}{Shotaro Akaho}, \bibinfo{person}{Hideki Asoh}, {and} \bibinfo{person}{Jun Sakuma}.} \bibinfo{year}{2012}\natexlab{}.
\newblock \showarticletitle{Fairness-aware classifier with prejudice remover regularizer}. In \bibinfo{booktitle}{\emph{Proceedings of the European Conference on Machine Learning and Knowledge Discovery in Databases, {ECML/PKDD} 2012}}. \bibinfo{pages}{35--50}.
\newblock


\bibitem[Kanei et~al\mbox{.}(2019)]%
        {kanei2019precise}
\bibfield{author}{\bibinfo{person}{Fumihiro Kanei}, \bibinfo{person}{Daiki Chiba}, \bibinfo{person}{Kunio Hato}, {and} \bibinfo{person}{Mitsuaki Akiyama}.} \bibinfo{year}{2019}\natexlab{}.
\newblock \showarticletitle{Precise and robust detection of advertising fraud}. In \bibinfo{booktitle}{\emph{Proceedings of the 2019 IEEE 43rd Annual Computer Software and Applications Conference, COMPSAC 2019}}. \bibinfo{pages}{776--785}.
\newblock


\bibitem[Kitchenham et~al\mbox{.}(2017)]%
        {eseKitchenhamMBKBC17}
\bibfield{author}{\bibinfo{person}{Barbara~A. Kitchenham}, \bibinfo{person}{Lech Madeyski}, \bibinfo{person}{David Budgen}, \bibinfo{person}{Jacky Keung}, \bibinfo{person}{Pearl Brereton}, \bibinfo{person}{Stuart~M. Charters}, \bibinfo{person}{Shirley Gibbs}, {and} \bibinfo{person}{Amnart Pohthong}.} \bibinfo{year}{2017}\natexlab{}.
\newblock \showarticletitle{Robust statistical methods for empirical software engineering}.
\newblock \bibinfo{journal}{\emph{Empirical Software Engineering}} \bibinfo{volume}{22}, \bibinfo{number}{2} (\bibinfo{year}{2017}), \bibinfo{pages}{579--630}.
\newblock


\bibitem[Kotsiantis et~al\mbox{.}(2006)]%
        {kotsiantis2006handling}
\bibfield{author}{\bibinfo{person}{Sotiris Kotsiantis}, \bibinfo{person}{Dimitris Kanellopoulos}, \bibinfo{person}{Panayiotis Pintelas}, {et~al\mbox{.}}} \bibinfo{year}{2006}\natexlab{}.
\newblock \showarticletitle{Handling imbalanced datasets: A review}.
\newblock \bibinfo{journal}{\emph{GESTS International Transactions on Computer Science and Engineering}} \bibinfo{volume}{30}, \bibinfo{number}{1} (\bibinfo{year}{2006}), \bibinfo{pages}{25--36}.
\newblock


\bibitem[Li et~al\mbox{.}(2023)]%
        {corrabs230802935}
\bibfield{author}{\bibinfo{person}{Xinyue Li}, \bibinfo{person}{Zhenpeng Chen}, \bibinfo{person}{Jie~M. Zhang}, \bibinfo{person}{Federica Sarro}, \bibinfo{person}{Ying Zhang}, {and} \bibinfo{person}{Xuanzhe Liu}.} \bibinfo{year}{2023}\natexlab{}.
\newblock \showarticletitle{Dark-skin individuals are at more risk on the street: Unmasking fairness issues of autonomous driving systems}.
\newblock \bibinfo{journal}{\emph{CoRR}}  \bibinfo{volume}{abs/2308.02935} (\bibinfo{year}{2023}).
\newblock


\bibitem[Li et~al\mbox{.}(2022)]%
        {icseLiMC0WZX22}
\bibfield{author}{\bibinfo{person}{Yanhui Li}, \bibinfo{person}{Linghan Meng}, \bibinfo{person}{Lin Chen}, \bibinfo{person}{Li Yu}, \bibinfo{person}{Di Wu}, \bibinfo{person}{Yuming Zhou}, {and} \bibinfo{person}{Baowen Xu}.} \bibinfo{year}{2022}\natexlab{}.
\newblock \showarticletitle{Training data debugging for the fairness of machine learning software}. In \bibinfo{booktitle}{\emph{Proceedings of the 44th {IEEE/ACM} 44th International Conference on Software Engineering, {ICSE} 2022}}. \bibinfo{pages}{2215--2227}.
\newblock


\bibitem[Mann and Whitney(1947)]%
        {mann1947test}
\bibfield{author}{\bibinfo{person}{Henry~B Mann} {and} \bibinfo{person}{Donald~R Whitney}.} \bibinfo{year}{1947}\natexlab{}.
\newblock \showarticletitle{On a test of whether one of two random variables is stochastically larger than the other}.
\newblock \bibinfo{journal}{\emph{The Annals of Mathematical Statistics}} (\bibinfo{year}{1947}), \bibinfo{pages}{50--60}.
\newblock


\bibitem[Mehrabi et~al\mbox{.}(2021)]%
        {csurMehrabiMSLG21}
\bibfield{author}{\bibinfo{person}{Ninareh Mehrabi}, \bibinfo{person}{Fred Morstatter}, \bibinfo{person}{Nripsuta Saxena}, \bibinfo{person}{Kristina Lerman}, {and} \bibinfo{person}{Aram Galstyan}.} \bibinfo{year}{2021}\natexlab{}.
\newblock \showarticletitle{A survey on bias and fairness in machine learning}.
\newblock \bibinfo{journal}{\emph{{ACM Computing Surveys}}} \bibinfo{volume}{54}, \bibinfo{number}{6} (\bibinfo{year}{2021}), \bibinfo{pages}{115:1--115:35}.
\newblock


\bibitem[Moussa and Sarro(2022)]%
        {isstaMoussaS22}
\bibfield{author}{\bibinfo{person}{Rebecca Moussa} {and} \bibinfo{person}{Federica Sarro}.} \bibinfo{year}{2022}\natexlab{}.
\newblock \showarticletitle{On the use of evaluation measures for defect prediction studies}. In \bibinfo{booktitle}{\emph{Proceedings of the 31st {ACM} {SIGSOFT} International Symposium on Software Testing and Analysis, ISSTA 2022}}. \bibinfo{pages}{101--113}.
\newblock


\bibitem[Myers et~al\mbox{.}(2013)]%
        {myers2013research}
\bibfield{author}{\bibinfo{person}{Jerome~L Myers}, \bibinfo{person}{Arnold~D Well}, {and} \bibinfo{person}{Robert~F Lorch~Jr}.} \bibinfo{year}{2013}\natexlab{}.
\newblock \bibinfo{booktitle}{\emph{Research design and statistical analysis}}.
\newblock \bibinfo{publisher}{Routledge}.
\newblock


\bibitem[Peng et~al\mbox{.}(2023)]%
        {fairmaskpaper}
\bibfield{author}{\bibinfo{person}{Kewen Peng}, \bibinfo{person}{Joymallya Chakraborty}, {and} \bibinfo{person}{Tim Menzies}.} \bibinfo{year}{2023}\natexlab{}.
\newblock \showarticletitle{FairMask: Better Fairness via Model-Based Rebalancing of Protected Attributes}.
\newblock \bibinfo{journal}{\emph{{IEEE} Transactions on Software Engineering}} \bibinfo{volume}{49}, \bibinfo{number}{4} (\bibinfo{year}{2023}), \bibinfo{pages}{2426--2439}.
\newblock


\bibitem[Pleiss et~al\mbox{.}(2017)]%
        {COpaper}
\bibfield{author}{\bibinfo{person}{Geoff Pleiss}, \bibinfo{person}{Manish Raghavan}, \bibinfo{person}{Felix Wu}, \bibinfo{person}{Jon~M. Kleinberg}, {and} \bibinfo{person}{Kilian~Q. Weinberger}.} \bibinfo{year}{2017}\natexlab{}.
\newblock \showarticletitle{On fairness and calibration}. In \bibinfo{booktitle}{\emph{Proceedings of the Annual Conference on Neural Information Processing Systems 2017, NIPS 2017}}. \bibinfo{pages}{5680--5689}.
\newblock


\bibitem[Ramyachitra and Manikandan(2014)]%
        {ramyachitra2014imbalanced}
\bibfield{author}{\bibinfo{person}{D Ramyachitra} {and} \bibinfo{person}{Parasuraman Manikandan}.} \bibinfo{year}{2014}\natexlab{}.
\newblock \showarticletitle{Imbalanced dataset classification and solutions: A review}.
\newblock \bibinfo{journal}{\emph{International Journal of Computing and Business Research}} \bibinfo{volume}{5}, \bibinfo{number}{4} (\bibinfo{year}{2014}), \bibinfo{pages}{1--29}.
\newblock


\bibitem[Sarro(2023)]%
        {SarroRE23}
\bibfield{author}{\bibinfo{person}{Federica Sarro}.} \bibinfo{year}{2023}\natexlab{}.
\newblock \showarticletitle{Search-based software engineering in the era of modern software systems}. In \bibinfo{booktitle}{\emph{Proceedings of the 31st {IEEE} International Requirements Engineering Conference, {RE} 2023}}.
\newblock


\bibitem[Soremekun et~al\mbox{.}(2022)]%
        {soremekun2022software}
\bibfield{author}{\bibinfo{person}{Ezekiel Soremekun}, \bibinfo{person}{Mike Papadakis}, \bibinfo{person}{Maxime Cordy}, {and} \bibinfo{person}{Yves~Le Traon}.} \bibinfo{year}{2022}\natexlab{}.
\newblock \showarticletitle{Software fairness: An analysis and survey}.
\newblock \bibinfo{journal}{\emph{arXiv preprint arXiv:2205.08809}} (\bibinfo{year}{2022}).
\newblock


\bibitem[Tao et~al\mbox{.}(2022)]%
        {sigsoftTaoSHF022}
\bibfield{author}{\bibinfo{person}{Guanhong Tao}, \bibinfo{person}{Weisong Sun}, \bibinfo{person}{Tingxu Han}, \bibinfo{person}{Chunrong Fang}, {and} \bibinfo{person}{Xiangyu Zhang}.} \bibinfo{year}{2022}\natexlab{}.
\newblock \showarticletitle{{RULER:} Discriminative and iterative adversarial training for deep neural network fairness}. In \bibinfo{booktitle}{\emph{Proceedings of the 30th {ACM} Joint European Software Engineering Conference and Symposium on the Foundations of Software Engineering, {ESEC/FSE} 2022}}. \bibinfo{pages}{1173--1184}.
\newblock


\bibitem[Vargha and Delaney(2000)]%
        {vargha2000critique}
\bibfield{author}{\bibinfo{person}{Andr{\'a}s Vargha} {and} \bibinfo{person}{Harold~D Delaney}.} \bibinfo{year}{2000}\natexlab{}.
\newblock \showarticletitle{A critique and improvement of the CL common language effect size statistics of McGraw and Wong}.
\newblock \bibinfo{journal}{\emph{Journal of Educational and Behavioral Statistics}} \bibinfo{volume}{25}, \bibinfo{number}{2} (\bibinfo{year}{2000}), \bibinfo{pages}{101--132}.
\newblock


\bibitem[Wick et~al\mbox{.}(2019)]%
        {nipsWickpT19}
\bibfield{author}{\bibinfo{person}{Michael~L. Wick}, \bibinfo{person}{Swetasudha Panda}, {and} \bibinfo{person}{Jean{-}Baptiste Tristan}.} \bibinfo{year}{2019}\natexlab{}.
\newblock \showarticletitle{Unlocking fairness: A trade-off revisited}. In \bibinfo{booktitle}{\emph{Proceedings of the Annual Conference on Neural Information Processing Systems 2019, NeurIPS 2019}}. \bibinfo{pages}{8780--8789}.
\newblock


\bibitem[Zafar et~al\mbox{.}(2017)]%
        {aistatsZafarVGG17}
\bibfield{author}{\bibinfo{person}{Muhammad~Bilal Zafar}, \bibinfo{person}{Isabel Valera}, \bibinfo{person}{Manuel Gomez{-}Rodriguez}, {and} \bibinfo{person}{Krishna~P. Gummadi}.} \bibinfo{year}{2017}\natexlab{}.
\newblock \showarticletitle{Fairness constraints: Mechanisms for fair classification}. In \bibinfo{booktitle}{\emph{Proceedings of the 20th International Conference on Artificial Intelligence and Statistics, {AISTATS} 2017}}. \bibinfo{pages}{962--970}.
\newblock


\bibitem[Zhang et~al\mbox{.}(2018)]%
        {ADVpaper}
\bibfield{author}{\bibinfo{person}{Brian~Hu Zhang}, \bibinfo{person}{Blake Lemoine}, {and} \bibinfo{person}{Margaret Mitchell}.} \bibinfo{year}{2018}\natexlab{}.
\newblock \showarticletitle{Mitigating unwanted biases with adversarial learning}. In \bibinfo{booktitle}{\emph{Proceedings of the 2018 {AAAI/ACM} Conference on AI, Ethics, and Society, {AIES} 2018}}. \bibinfo{pages}{335--340}.
\newblock


\bibitem[Zhang and Harman(2021)]%
        {icseZhangH21}
\bibfield{author}{\bibinfo{person}{Jie~M. Zhang} {and} \bibinfo{person}{Mark Harman}.} \bibinfo{year}{2021}\natexlab{}.
\newblock \showarticletitle{Ignorance and prejudice in software fairness}. In \bibinfo{booktitle}{\emph{Proceedings of the 43rd {IEEE/ACM} International Conference on Software Engineering, {ICSE} 2021}}. \bibinfo{pages}{1436--1447}.
\newblock


\bibitem[Zhang et~al\mbox{.}(2022)]%
        {jieMLsurvey}
\bibfield{author}{\bibinfo{person}{Jie~M. Zhang}, \bibinfo{person}{Mark Harman}, \bibinfo{person}{Lei Ma}, {and} \bibinfo{person}{Yang Liu}.} \bibinfo{year}{2022}\natexlab{}.
\newblock \showarticletitle{Machine learning testing: Survey, landscapes and horizons}.
\newblock \bibinfo{journal}{\emph{{IEEE} Transactions on Software Engineering}} \bibinfo{volume}{48}, \bibinfo{number}{2} (\bibinfo{year}{2022}), \bibinfo{pages}{1--36}.
\newblock


\bibitem[Zhang and Sun(2022)]%
        {sigsoftZhang022}
\bibfield{author}{\bibinfo{person}{Mengdi Zhang} {and} \bibinfo{person}{Jun Sun}.} \bibinfo{year}{2022}\natexlab{}.
\newblock \showarticletitle{Adaptive fairness improvement based on causality analysis}. In \bibinfo{booktitle}{\emph{Proceedings of the 30th {ACM} Joint European Software Engineering Conference and Symposium on the Foundations of Software Engineering, {ESEC/FSE} 2022}}. \bibinfo{pages}{6--17}.
\newblock


\bibitem[Zhang et~al\mbox{.}(2020)]%
        {icseZhangW0D0WDD20}
\bibfield{author}{\bibinfo{person}{Peixin Zhang}, \bibinfo{person}{Jingyi Wang}, \bibinfo{person}{Jun Sun}, \bibinfo{person}{Guoliang Dong}, \bibinfo{person}{Xinyu Wang}, \bibinfo{person}{Xingen Wang}, \bibinfo{person}{Jin~Song Dong}, {and} \bibinfo{person}{Ting Dai}.} \bibinfo{year}{2020}\natexlab{}.
\newblock \showarticletitle{White-box fairness testing through adversarial sampling}. In \bibinfo{booktitle}{\emph{Proceedings of the 42nd International Conference on Software Engineering, ICSE 2020}}. \bibinfo{pages}{949--960}.
\newblock


\bibitem[Zheng et~al\mbox{.}(2022)]%
        {icseZhengCD0CJW0C22}
\bibfield{author}{\bibinfo{person}{Haibin Zheng}, \bibinfo{person}{Zhiqing Chen}, \bibinfo{person}{Tianyu Du}, \bibinfo{person}{Xuhong Zhang}, \bibinfo{person}{Yao Cheng}, \bibinfo{person}{Shouling Ji}, \bibinfo{person}{Jingyi Wang}, \bibinfo{person}{Yue Yu}, {and} \bibinfo{person}{Jinyin Chen}.} \bibinfo{year}{2022}\natexlab{}.
\newblock \showarticletitle{NeuronFair: Interpretable white-box fairness testing through biased neuron identification}. In \bibinfo{booktitle}{\emph{Proceedings of the 44th {IEEE/ACM} 44th International Conference on Software Engineering, {ICSE} 2022}}. \bibinfo{pages}{1519--1531}.
\newblock


\end{thebibliography}

\end{document}